\documentclass[10pt,twocolumn,letterpaper]{article}

\usepackage{cvpr}      

\definecolor{cvprblue}{rgb}{0.21,0.49,0.74}
\usepackage[pagebackref,breaklinks,colorlinks,allcolors=cvprblue]{hyperref}
\usepackage{float}
\usepackage{pifont}
\usepackage{caption}
\usepackage{multirow}
\usepackage{listings}
\usepackage{xcolor}
\usepackage{minitoc}
\lstdefinelanguage{json}{
    basicstyle=\ttfamily\footnotesize,
    numbers=none,
    numberstyle=\tiny\color{gray},
    stepnumber=1,
    numbersep=5pt,
    showstringspaces=false,
    breaklines=true,
    frame=single,
    backgroundcolor=\color{gray!10},
    literate=
     *{0}{{{\color{blue}0}}}{1}
      {1}{{{\color{blue}1}}}{1}
      {2}{{{\color{blue}2}}}{1}
      {3}{{{\color{blue}3}}}{1}
      {4}{{{\color{blue}4}}}{1}
      {5}{{{\color{blue}5}}}{1}
      {6}{{{\color{blue}6}}}{1}
      {7}{{{\color{blue}7}}}{1}
      {8}{{{\color{blue}8}}}{1}
      {9}{{{\color{blue}9}}}{1}
      {:}{{{\color{black}:}}}{1}
      {,}{{{\color{black},}}}{1}
      {"}{{{\color{red}"}}}{1},
}

\title{\textbf{Plot'n Polish}: Zero-shot Story Visualization and Disentangled Editing with Text-to-Image Diffusion Models}

\author{Kiymet Akdemir$^{1}$ \quad Jing Shi$^{2}$ \quad Kushal Kafle$^{2}$ \quad Brian Price$^{2}$ \quad Pinar Yanardag$^{1}$ \vspace{0.3em} \\
{\normalsize $^1$Virginia Tech} \quad
{\normalsize $^2$Adobe Research} \\
\small{\url{plotnpolish.github.io}}
}

\begin{document}
\twocolumn[{
\maketitle
\begin{center}
    \captionsetup{type=figure}
    \vspace{-1em}
\newcommand{\imwidth}{1\textwidth}

\begin{tabular}{@{}c@{}}
 
\parbox{\imwidth}{\centering\includegraphics[width=0.8\textwidth]{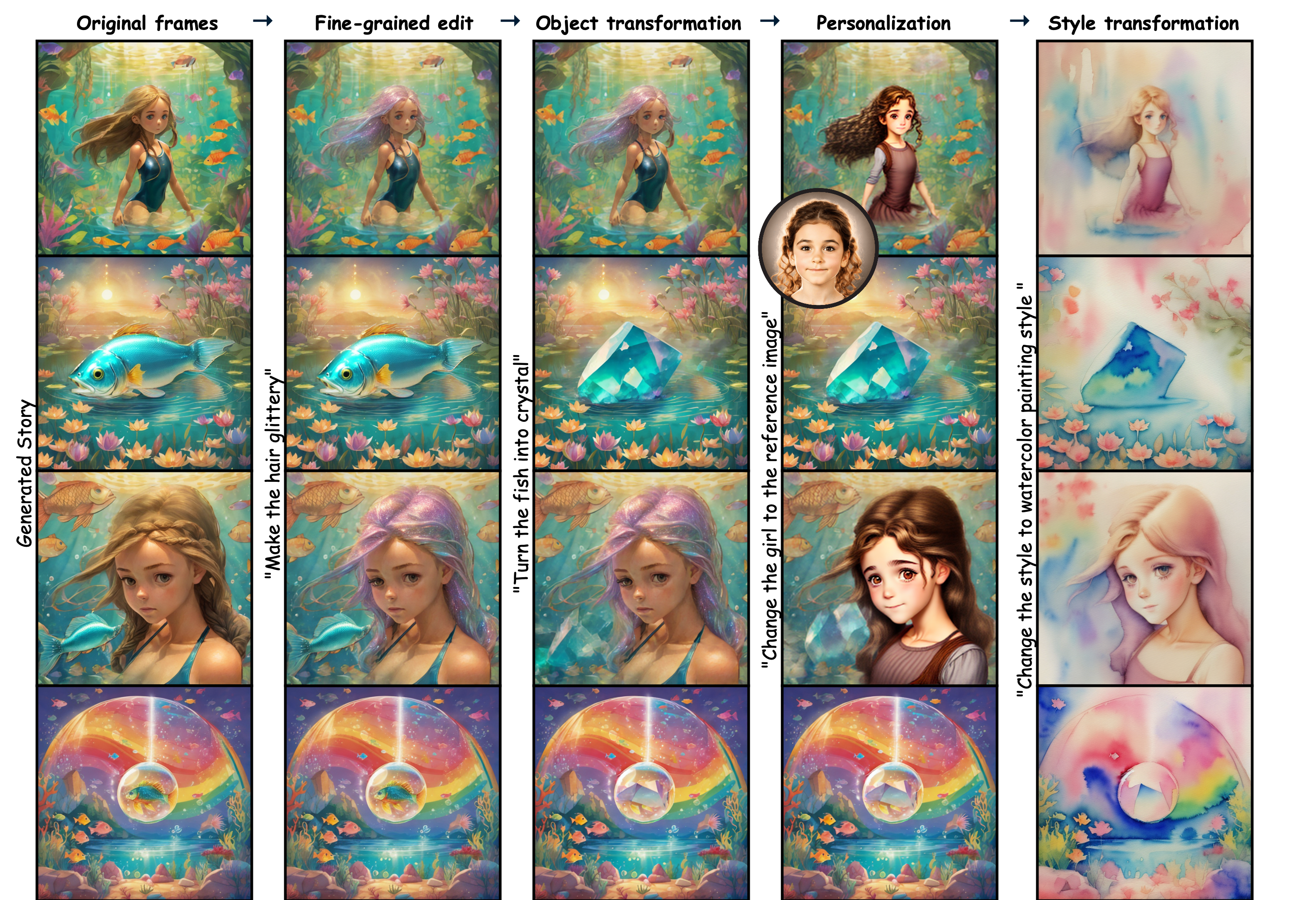}}
\\
\vspace{1em}
\end{tabular}
    \vspace{-3em}
    \captionof{figure}{We introduce \texttt{Plot'n Polish}, a training-free approach for creating and refining story visualizations. Our framework enables users to adjust story elements through fine or coarse-grained edits. Users can alter elements like hairstyles or clothing, transform objects or styles, and customize characters iteratively, all directed through text prompts and without the need for any manual intervention.}
    \label{fig:teaser}
\end{center}
}] 
\begin{abstract}
\label{sec:abstract}
Text-to-image diffusion models have demonstrated significant capabilities to generate diverse and detailed visuals in various domains, and story visualization is emerging as a particularly promising application. However, as their use in real-world creative domains increases, the need for providing enhanced control, refinement, and the ability to modify images post-generation in a consistent manner becomes an important challenge. Existing methods often lack the flexibility to apply fine or coarse edits while maintaining visual and narrative consistency across multiple frames, preventing creators from seamlessly crafting and refining their visual stories. To address these challenges, we introduce Plot'n Polish, a zero-shot framework that enables consistent story generation and provides fine-grained control over story visualizations at various levels of detail.
\end{abstract}    
\section{Introduction}
\label{sec:intro}

Text-to-image diffusion models have emerged as powerful tools for generating high-quality, detailed, and diverse images across various domains \cite{rombach2022high, ddpm, esser2024scaling}. Story visualization is a particularly exciting application, aiming to generate image sequences that form a coherent narrative guided by sequential text prompts. However, as these models are increasingly used in creative processes, there's a significant challenge in providing not only coherent story sequences but also enhanced control and refinement. Storytelling often requires the ability to create and modify the narrative, a flexibility current methods fail to address. Creators might want to alter the storyline by introducing new plot elements or characters during generation. Additionally, after story visuals are generated, creators may wish to make fine-grained adjustments—such as adding accessories to a character or changing attire colors or more substantial changes like replacing a character or altering the visual style of the story.

Current approaches to story visualization face several key limitations (see Table~\ref{tab:compare_methods}), and lack flexibility for post-generation edits, forcing users to regenerate entire sequences to make changes. Many methods demand extensive, computationally expensive training on large datasets \cite{tao2024storyimager}, making them less accessible to a broad range of users. Additionally, several existing techniques generate each story frame independently, leading to visual inconsistencies, particularly when characters, objects, or settings need to evolve fluidly across multiple scenes \cite{maharana2022storydall, pan2024synthesizing, rahman2023make}. Moreover, methods that generate all frames simultaneously, although maintaining coherence, often produce images that are very similar to each other \cite{zhou2024storydiffusion}. Some approaches incorporate sketches or personalization to introduce control, but these methods often limit versatility \cite{cheng2024autostudio, wang2023autostory}. Although some approaches explored  editing within the context of story visualization, their editing capabilities are limited to individual frames and do not support consistent, multi-frame modifications across the entire narrative which is crucial for story coherence \cite{cheng2024autostudio,cheng2024theatergen}.

% While some models \cite{wang2023autostory} employ personalization techniques to fine-tune specific characteristics, this often restricts creative diversity, limiting the exploration of novel visual directions.

To address these limitations, we present \texttt{Plot'n Polish}, a novel zero-shot framework that provides users with comprehensive control over creating and refining story visualizations while maintaining consistency across multiple frames. Our method enables users to make both fine-grained adjustments and substantial modifications, such as replacing a character or applying stylistic changes across all scenes (Fig.~\ref{fig:teaser}) without any training or fine-tuning. This flexibility allows for experimentation with different artistic styles and visual elements without the need to regenerate frames individually. Unlike previous methods that lack post-editing flexibility, our approach incorporates a novel editing mechanism that leverages inter-frame correspondences, ensuring edits are consistently applied across multiple frames. Our method supports both the generation of consistent story visuals and the seamless editing of generated frames, as well as user-provided story visuals, whether created by other methods or sourced from published books. By maintaining both narrative continuity and stylistic coherence throughout the sequence, \texttt{Plot'n Polish} empowers creators to  craft, refine, and enhance their visual narratives.  Our contributions can be summarized as follows:

\begin{itemize}
    \item We introduce \texttt{Plot’n Polish}, a novel method that supports both initial story visualization and post generation modifications that ensures multi-frame consistency for both local and global edits.
    \item Our approach seamlessly integrates with existing workflows by supporting the editing of previously generated or user-provided story visuals, including real-world illustrations from published storybooks. 
    \item Our method accommodates a wide range of editing tasks, from fine-grained adjustments to character transformations and personalization, empowering users to refine story visuals with precision, adapt elements to fit specific narratives, and seamlessly integrate customized details into the story frames.
    \item Through comprehensive experiments, we show that Plot’n Polish outperforms state-of-the-art story visualization and editing methods in terms of consistency, text alignment, and editing flexibility.
\end{itemize}

\begin{table}
    \centering
\caption{Plot'n Polish is the only story visualization method that enables multi-frame editing.}
\resizebox{0.5\textwidth}{!}{
\begin{tabular}{lcccccc}
\toprule
 Methods/ task  & Multi-turn Editing & Multi-frame Editing & Publicly Available\\ 
\midrule
AutoStory~\cite{wang2023autostory}  & \ding{55}  & \ding{55} & \ding{55}\\
TaleCrafter~\cite{gong2023talecrafter}  & \ding{55}  & \ding{55} & \ding{55}\\
Intelligent Grimm~\cite{liu2024intelligent}  & \ding{55} & \ding{55} & \checkmark\\
StoryDiffusion~\cite{zhou2024storydiffusion} & \ding{55} & \ding{55} & \checkmark \\
ConsiStory~\cite{tewel2024consistory} & \ding{55} & \ding{55} & \checkmark \\
Theatergen~\cite{cheng2024theatergen}  & \checkmark & \ding{55} & \checkmark \\
AutoStudio \cite{cheng2024autostudio} & \checkmark & \ding{55} & \checkmark \\
\textbf{Plot'n Polish (ours)}  & \checkmark & \checkmark & \checkmark \\
\bottomrule
\end{tabular}
  }
\label{tab:compare_methods}
\end{table}

% However, it is important to note that while this component generates consistent narratives, using these narratives with standard T2I models often leads to inconsistent visualizations, as illustrated in Fig. \ref{fig:motivation}. To address this issue, we introduce a novel framework capable of producing consistent story visualizations. Additionally, this framework allows for refinements based on user prompts if desired.
\section{Related Work}
\label{sec:related-work}

\paragraph{Story Visualization} Recent work has explored the use of text-to-image models for story generation.
\cite{pan2024synthesizing, liu2024intelligent, rahman2023make, yang2024seedstory} introduce autoregressive models that require training on specific datasets, limiting their generalization capabilities and compromising image quality compared to the base model.
\cite{gong2023talecrafter, wang2023autostory, avrahami2023chosen} rely on LoRA~\cite{hu2021lora} models, requiring finetuning for each character. \cite{jeong2023zeroshotgenerationcoherentstorybook} focuses on face editing using a personalization method.
\cite{tao2024storyimager} generates multi-frame storyboards but lacks fine-grained editing capabilities.
\cite{cheng2024autostudio} introduces an additional U-Net~\cite{unet} to improve identity consistency. While it addresses multi-turn editing, it does not ensure consistency across multiple edited images—an essential aspect for interactive storytelling.  While methods such as 
\cite{liu2025onepromptonestoryfreelunchconsistenttexttoimage, zhou2024storydiffusion, tewel2024consistory} utilize attention sharing for story generation, they do not perform editing. Overall, our method is the only method that supports multi-frame editing in the story visualization context, allowing users to apply consistent edits across the entire story sequence.

\paragraph{Image Editing with Diffusion Models.} Many diffusion-based editing approaches rely on text prompts to specify modifications.
\cite{instructpix2pix} enables text-driven image edits but often results in entangled edits, affecting unintended areas.
Some works improve control over the editing process~\cite{hertz2022prompt, zhang2023adding}, such as ControlNet~\cite{zhang2023adding}, which uses conditional inputs to adjust specific image attributes. Similarly, \cite{unitune} ensures content preservation by fine-tuning the diffusion model on the input image.
\cite{Tumanyan_2023_CVPR} retrieves inversion noise and applies denoising for feature reconstruction. To enhance flexibility, recent approaches such as~\cite{brack2023sega, brack2024ledits, liu2022compositional} decompose edits into multiple components, enabling finer control over modifications.   However, most of these methods focus on editing single images, overlooking multi-frame consistency. In contrast, our method edits the same object across multiple frames simultaneously by leveraging attention sharing for localized masked-area editing. 
 
\paragraph{Grid Priors.} Several works, including RAVE \cite{kara2024rave} and NeRFiller \cite{weber2023nerfillercompletingscenesgenerative}, utilize grid priors to achieve their goals, but our approach differs significantly in its application and fine-grained control. While RAVE~\cite{kara2024rave} uses a grid-based latent representation to achieve temporal consistency for edits on existing videos, it does not provide disentangled editing; applying edits to a character can often lead to unintended changes in the background or other scene elements. In contrast, our method is designed for disentangled editing and latent blending, ensuring that edits are applied in a localized and isolated manner within story frames. This enables us to create coherent narratives from scratch without affecting the overall scene. Similarly, StoryImager \cite{tao2024storyimager} generates a complete storyboard in a grid, which fundamentally restricts the number of frames and requires extensive training. Our approach, on the contrary, is training-free, enabling the generation of longer stories with on-the-fly editing capabilities. A more recent method, ORACLE~\cite{akdemir2024oracle}, uses a grid trick to train a LoRA~\cite{hu2021lora} for identity consistency. However, our method operates in a training-free manner, enabling zero-shot story generation and editing of multi-character sequences without any additional fine-tuning.
\section{Methodology}
\label{sec:method}

We present Plot 'n Polish, an end-to-end interactive pipeline that provides users with comprehensive control over creating and refining story visualizations while maintaining consistency across multiple frames. Our framework is flexible enough to edit existing story frames provided by users or to help users generate new stories from simple ideas, such as 'Create me a story about a girl and a cat discovering their backyard.' For users looking to create story visualizations from scratch, we first employ a simple strategy by utilizing LLMs to generate story narratives (see Appendix). Alternatively, users can provide plot texts and image prompts for the story visualization. 

\begin{figure}
    \centering
    \includegraphics[width=0.7\linewidth]{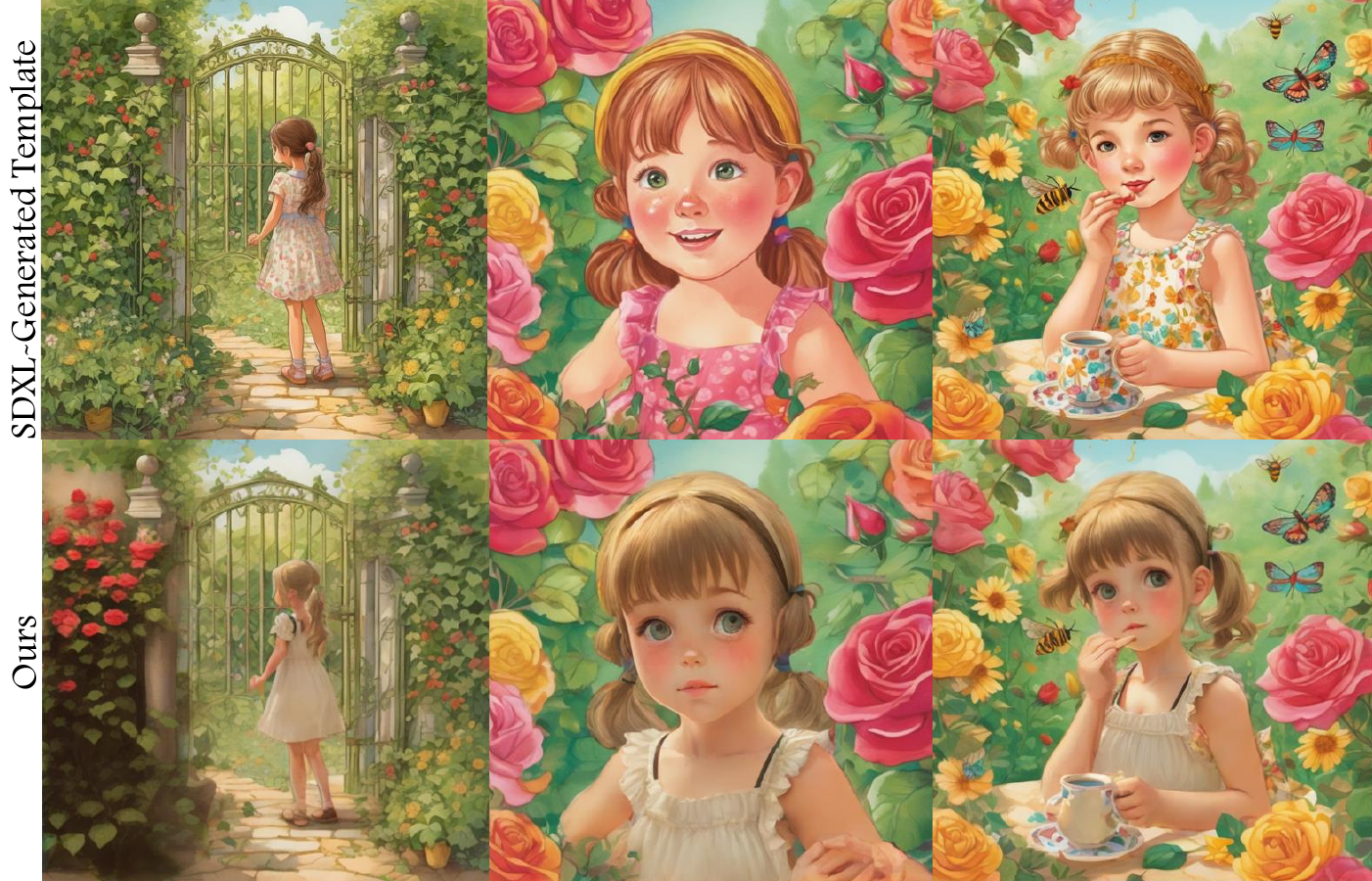}
    \vspace{-5pt}
    \caption{The story template is initially generated using an off-the-shelf T2I model, such as SDXL (top row). Our method edits these inconsistencies, transforming the template into a series of consistent story panels (bottom row).}
    \label{fig:motivation}
\end{figure}

\begin{figure*}[ht]
    \centering
    \includegraphics[width=0.80\linewidth]{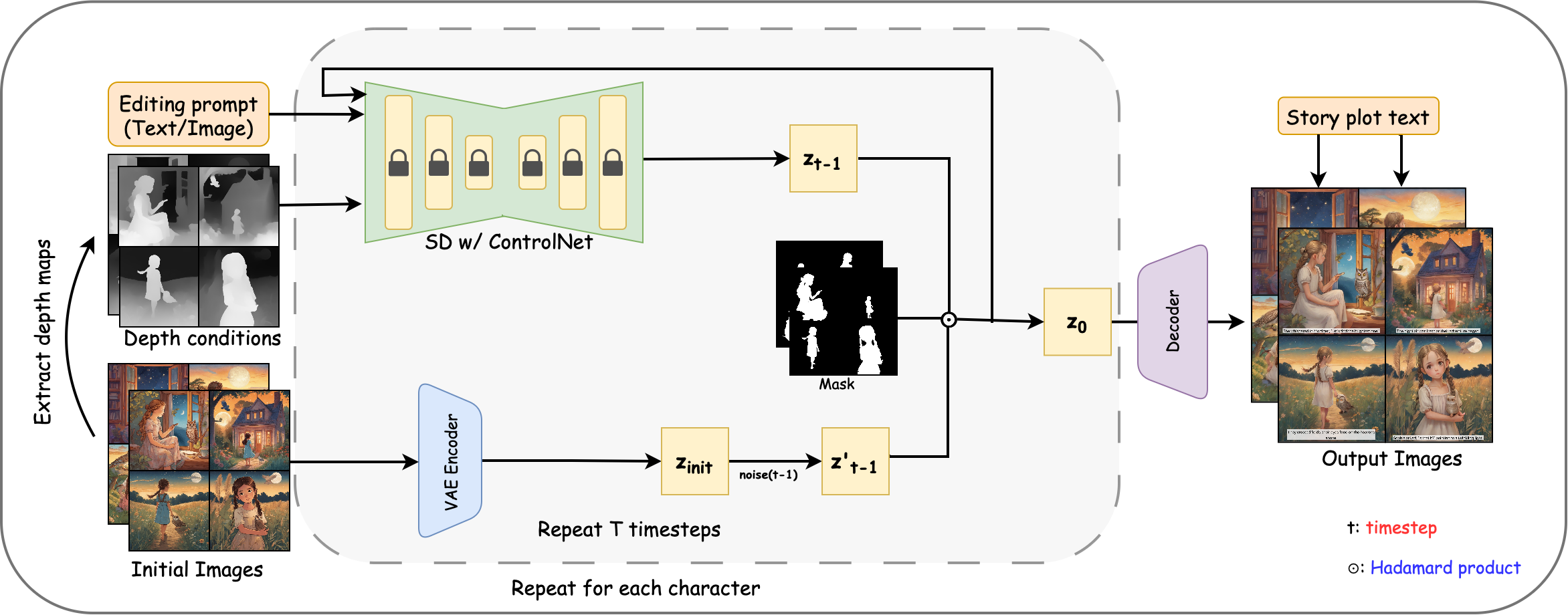}
    \caption{\textbf{An overview of Plot'n Polish.} Users can provide story plots for each frame and image prompts, or these can be generated by the LLM based on the story idea. The image prompts are used to create template images for visualizing the story. The editing framework takes editing prompts in the form of text or images, along with initial images to edit and extracted depth conditions.}

    \label{fig:method}
\end{figure*}
Our novel multi-frame editing method is designed for two main purposes: 1) ensuring consistency in initial story visualizations and 2) performing coherent edits based on user-provided text prompts. Both tasks are integrated into a unified framework with slight task-specific adaptations. See Figure~\ref{fig:method} for an overview of the pipeline.

\subsection{Story Visualization}
\label{sec:generation}

 Given a set of story prompts  \(\mathcal{P}\) either generated by an LLM or provided by the user,  we first create an initial story template of $n$ frames denoted as \(\mathcal{F} = \{F_1, F_2, \ldots, F_n\}\), corresponding to the prompt set \(\mathcal{P}\), using an off-the-shelf T2I model such as SDXL~\cite{podell2023sdxlimprovinglatentdiffusion}. This approach ensures that the diversity of layouts reflects the capabilities of the original model, including backgrounds and objects as described by the text prompt. However, the generated template often lacks consistency, with characters and details varying significantly across frames (refer to  Fig. \ref{fig:motivation} (top) where the girl character and her attire are inconsistent across different frames). To address this issue, we employ our novel multi-frame editing framework that refines the initial story template to achieve consistent character representation across frames (refer to Fig. \ref{fig:motivation} (bottom)) as described below. 

\subsection{Multi-Frame Editing}
\label{sec:edit}

Given the initial story template $\mathcal{F}$ generated in Section \ref{sec:generation}, a concept $k$ to modify (such as `boy') and editing prompt $\mathcal{P}_{edit}$ such as `a boy with short hair and striped t-shirt', we first extract masks \( \mathcal{M}_k = \{M_1, M_2, \dots, M_{n}\} \) using object detection and semantic segmentation models \cite{Cheng2024YOLOWorld, xiong2023efficientsamleveragedmaskedimage}. After obtaining the masks, we generate consistent images that will align with the editing prompt $\mathcal{P}_{edit}$ to modify the masked regions of the original frames while ensuring visual consistency. 
To enforce consistency across multiple frames, we leverage the grid prior \cite{weber2023nerfillercompletingscenesgenerative}, which enables interaction of spatial information across multiple frames. This approach ensures that modifications remain coherent throughout the sequence while allowing the model to leverage spatial relationships between frames. In addition, we incorporate ControlNet~\cite{controlnet} to preserve the structural integrity of the original image while guiding modifications within the specified regions.

Given \( n \) frames, we partition them into groups of size \( \gamma = \phi \times \beta\) and arrange them into a rectangular grid. The grid representation is defined as follows:

\[
    Grid(X_1, X_2, .... X_{\gamma}) = 
    \begin{bmatrix}
        X_1 & \dots & X_{\beta}\\
         \vdots & \ddots & \\
          X_{(\phi-1)*\beta+1} &  & X_{\gamma}
    \end{bmatrix}.
\]

We form grids \( z_{\text{grid}} \) corresponding to the latent representations, \( m_{\text{grid}} \) to the masks, \( d_{\text{grid}} \) to the depth conditions, and \( F_{\text{grid}} \) to the original frames, ensuring structured processing across multiple frames.
At each diffusion timestep \( t \in \{T, T-1, \dots, 1\} \), we update the latent representation for each grid as follows:

\[
z_{\text{grid}, t-1} \leftarrow \epsilon_\theta(z_{\text{grid}, t}, d_{\text{grid}}, t, \mathcal{P}_{edit}).
\]

To ensure that all frames interact with each other and to prevent isolated inconsistencies, we reform the grids at each timestep by randomly regrouping the frames. This process reduces memory overhead while providing cross-frame consistency.

To prevent modifications from unintentionally altering unmasked regions, we apply latent blending after each denoising step. Inspired by~\cite{blended_latent}, this approach ensures that the edited frames retain the original background details while applying changes only to the masked regions. We incorporate a secondary noised latent representation derived from the original grid, defined as:

\[
z'_{\text{grid}, t-1} = \sqrt{\alpha_{t-1}} \mathcal{E}(F_{\text{grid}}) + \sqrt{1 - \alpha_{t-1}} \epsilon,
\]

where \( \mathcal{E} \) denotes the encoder, \( \alpha_t \) controls the noise schedule, and \( \epsilon \sim \mathcal{N}(0, I) \) is a sampled Gaussian noise term. We then apply the latent blending process:

\[
z_{\text{grid}, t-1} \leftarrow z_{\text{grid}, t-1} \odot m + z'_{\text{grid}, t-1} \odot (1 - m),
\]

where \( m \) is the resized version of \( M_{\text{grid}} \) to match the latent resolution, and \( \odot \) denotes element-wise multiplication. For global edits such as style modifications, we omit this blending step, allowing changes to propagate across the entire image.

Once the denoising process is complete, we decode the final latent grid back into the pixel space to obtain the final images. This reconstruction step is performed as:

\[
x_{\text{grid}, 0} = \mathbf{D}(z_{\text{grid}, 0}),
\]

where \( \mathbf{D} \) represents the decoder of the diffusion model. By incorporating depth conditions and latent blending, we effectively balance consistency and diversity, ensuring that the final frames maintain their structural coherence while allowing for expressive modifications.

\paragraph{Personalization.} Our method enables personalization by allowing users to customize generated characters, animals, or objects using either a pre-trained LoRA~\cite{hu2021lora} model or a single reference image. The architecture of our pipeline is fundamentally compatible with the IP-Adapter \cite{ye2023ip-adapter}. By simply loading IP-Adapter's cross-attention layers, our model can process image-based prompts, seamlessly embedding personalized elements into the generation process.

\begin{figure*}[t!]
    \centering
    \includegraphics[width=\linewidth]{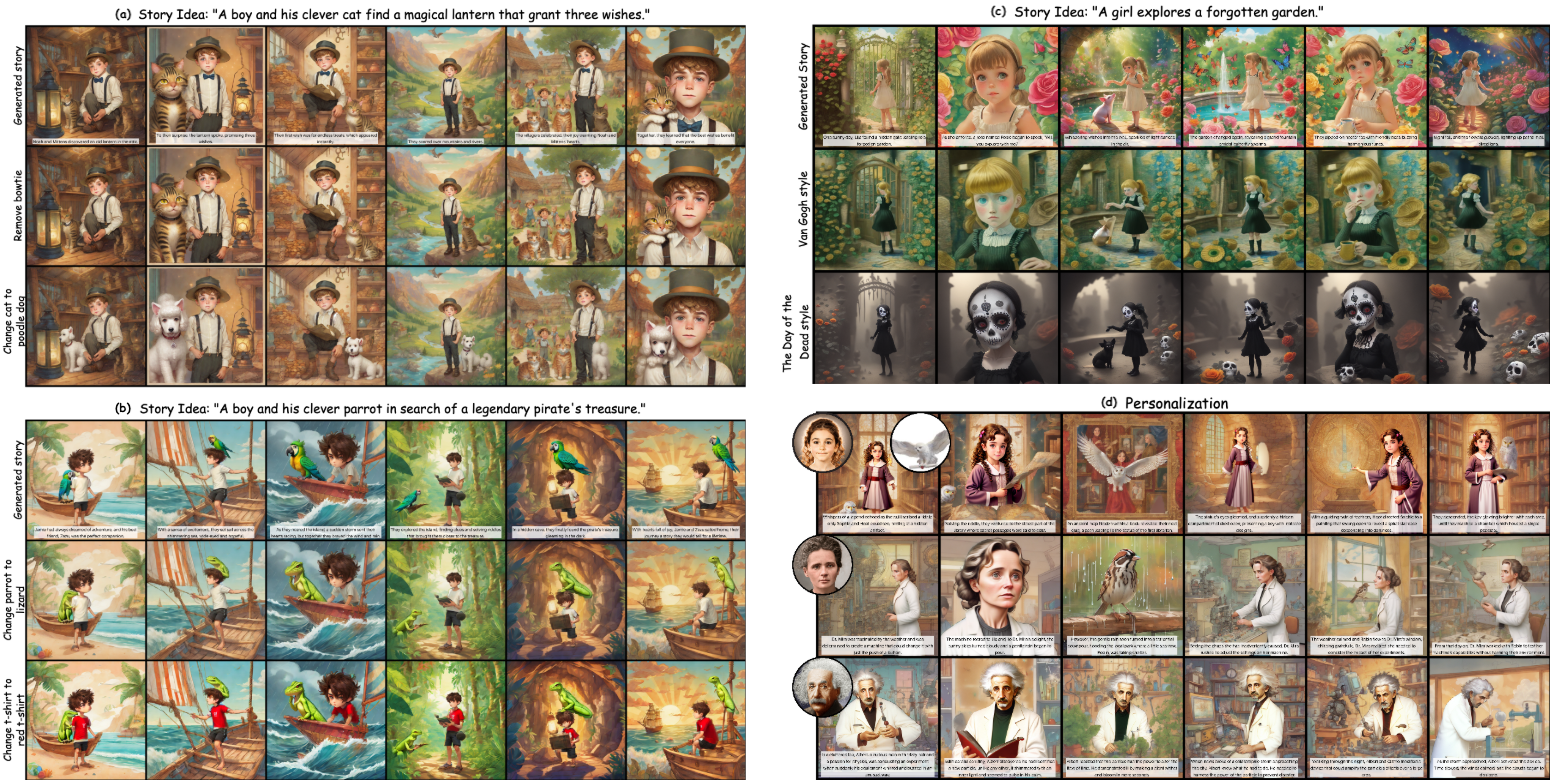}
    \caption{Qualitative results for Plot'n Polish. Our results demonstrate that Plot'n Polish excels in producing consistent visual narratives and allows for a wide range of successful edits including localized edits, character or object replacements, and personalization.}
    \label{fig:qual}
\end{figure*}
\section{Experiments}
\label{sec:exp}
We evaluated Plot'n Polish against both state-of-the-art story visualization methods~\cite{zhou2024storydiffusion, tewel2024consistory, cheng2024autostudio, liu2024intelligent} and editing methods~\cite{instructpix2pix,brack2024ledits,Tumanyan_2023_CVPR} through a series of qualitative and quantitative experiments. Our results demonstrate that Plot'n Polish excels in producing consistent visual stories and allows for a wide range of successful edits. Additionally, our method offers customization capabilities, enabling personalized storytelling experiences. 
\begin{figure*}[!t]
    \centering
    \includegraphics[width=0.8\textwidth]{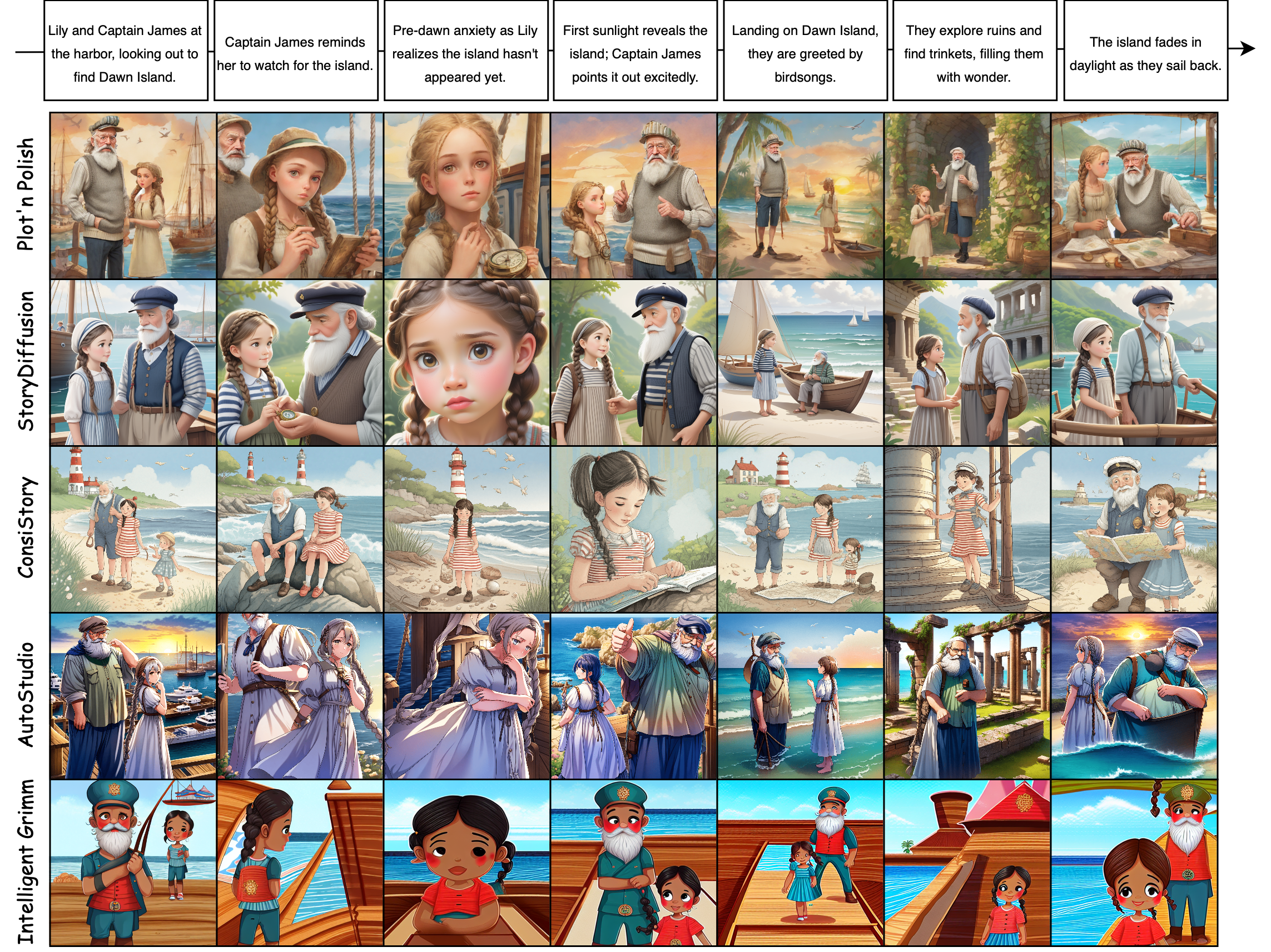}
    \caption{Qualitative comparison of our method with state-of-the-art story visualization methods, including StoryDiffusion, ConsiStory, AutoStudio, and Intelligent Grim. Our method outperforms competitors by maintaining consistent visual elements, such as attire and character features, across all panels, ensuring narrative coherence. In contrast, existing methods struggle with inconsistencies, blending errors, often breaking narrative flow and reducing clarity. }
    \label{fig:qual-comp}
\end{figure*}
\begin{figure}
    \centering
    \includegraphics[width=0.8\linewidth]{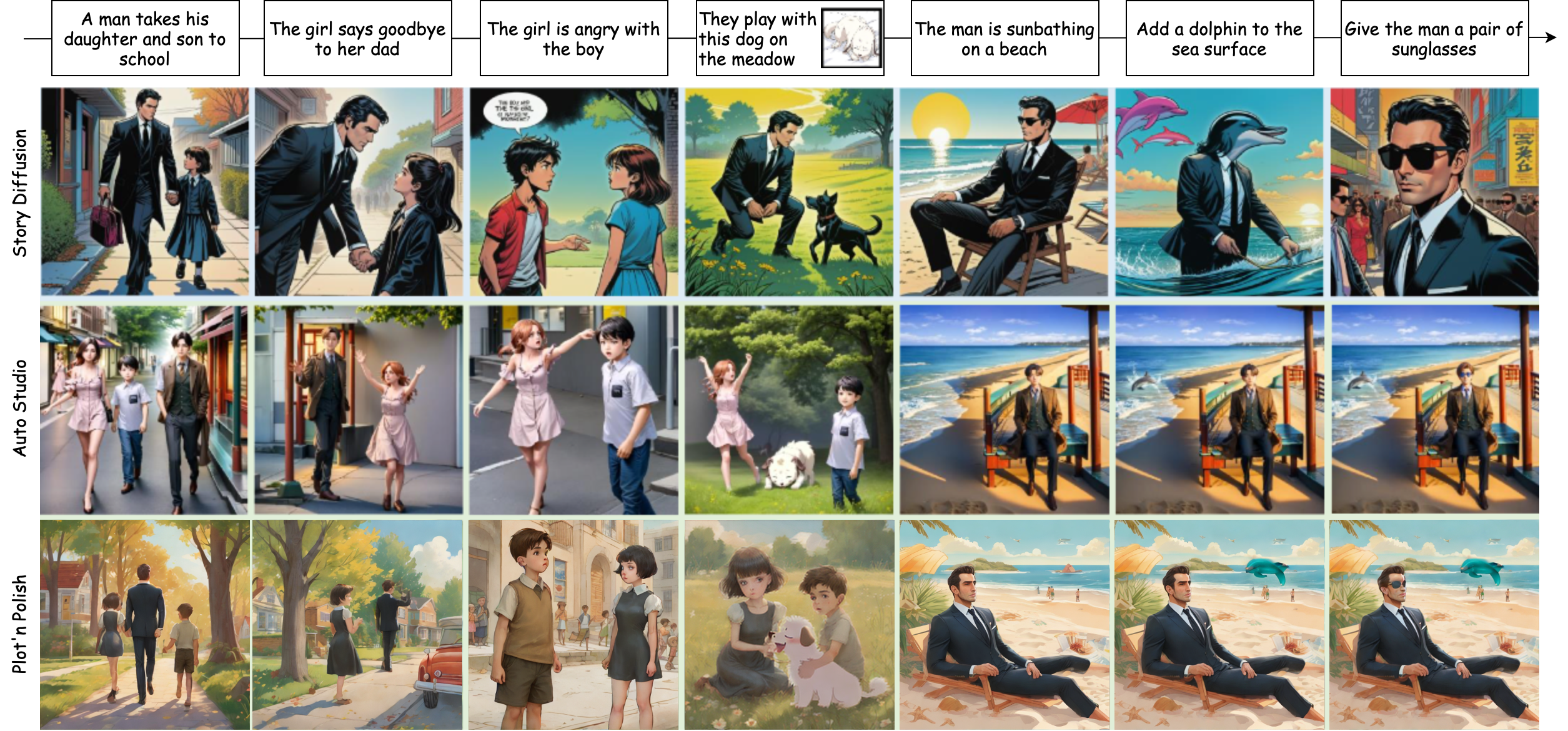}
    \caption{Qualitative Comparison for Editing. Our method outperforms StoryDiffusion and AutoStudio in accurately applying edits while maintaining narrative coherence. StoryDiffusion struggles with unintended changes and blending errors, such as merging a dolphin with the character, while AutoStudio misinterprets prompts, generating incorrect poses like having the character pose to the camera instead of sunbathing. In contrast, our method consistently captures the intent of the prompts, producing realistic and contextually appropriate edits.}
    \label{fig:edit-comp}
\end{figure}
\begin{figure}
    \centering
    \includegraphics[width=\linewidth]{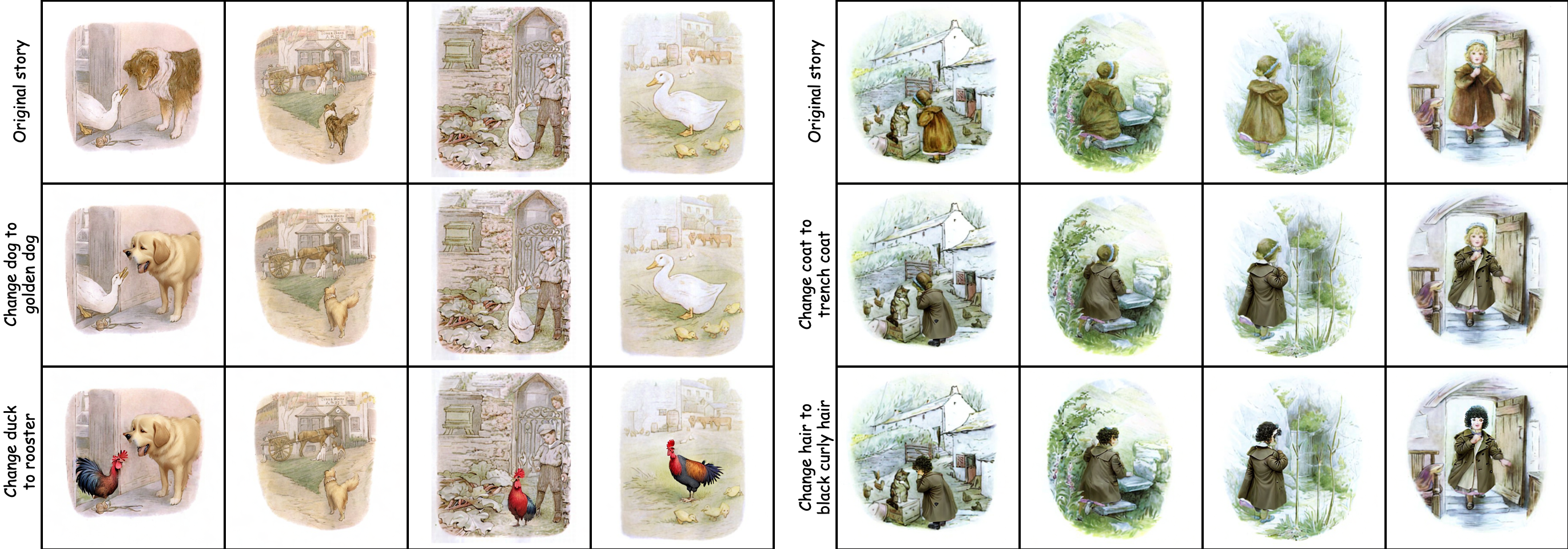}
    \caption{Our method can edit existing stories, ranging from transformations such as transforming characters (e.g., a duck to a rooster) to fine-grained modifications like altering clothing (e.g., a coat to a trench coat) or hairstyle.}
    \label{fig:existing}
\end{figure}
\begin{figure*}
    \centering
    \includegraphics[width=0.85\textwidth]{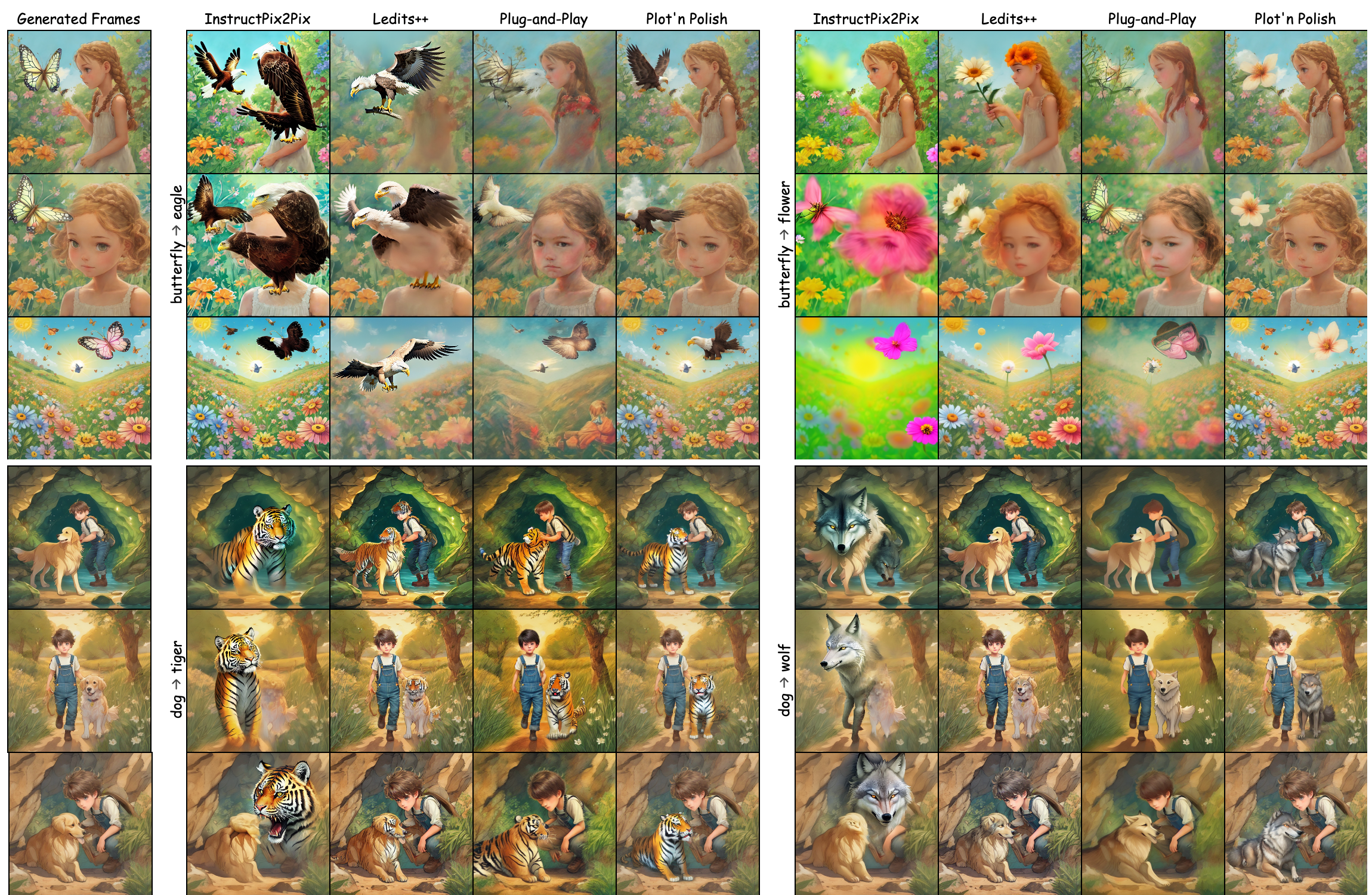}
    \caption{Qualitative comparison of editing methods. Our approach produces more \textbf{consistent} and \textbf{disentangled} edits, accurately applying transformations while preserving the background and character details. In contrast, LEDITS++~\cite{brack2024ledits} struggles with blending issues, as seen in the tiger edit where parts of the original dog remain. Plug-and-Play~\cite{Tumanyan_2023_CVPR} fails to generate certain objects, such as the eagle or the flower. InstructPix2Pix~\cite{instructpix2pix} often modifies unintended elements, like changing the girl’s hair color or distorting her face.}
    \label{fig:edit_comp}
\end{figure*}

\subsection{Experimental Setup}
All template images are generated using SDXL, while image editing is performed with Stable Diffusion (SD) 1.5. A 3×3 grid is employed for generating story panels. For additional editing examples using SDXL and Flux, refer to Supplementary Figure~\ref{fig:edit-xl}. For editing, we apply depth-conditioned ControlNet, using a depth condition of 0.4 for local edits and 1.0 for global edits. Experiments are conducted on a single NVIDIA A40 GPU.

\subsection{Qualitative Experiments}

\noindent \textbf{Qualitative Results.} Fig.~\ref{fig:qual} presents a diverse range of qualitative experiments demonstrating the versatility of our method for consistent story generation and editing based on simple text prompts. Our method is able to maintain characters consistency including retaining their attire, such as hats (Fig.~\ref{fig:qual}~(a)) or dresses (Fig.~\ref{fig:qual}~(c)), across multiple panels. Similarly, smaller objects like a fish (Fig.~\ref{fig:teaser}) or parrot (Fig.~\ref{fig:qual}~(b)) are accurately maintained throughout the story, demonstrating fine-grained visual and contextual coherence.

Moreover, our approach supports fine-grained edits, such as modifying specific object attributes like changing hair  to `glitter hair' (Fig. \ref{fig:teaser}), removing small objects like bowtie (Fig. \ref{fig:qual} (a)) or altering the color of a t-shirt to red (Fig. \ref{fig:qual} (b)). Additionally, our method enables character transformations, such as replacing a fish with a crystal (Fig. \ref{fig:teaser}) or changing a parrot into a lizard (Fig. \ref{fig:qual} (b)). These capabilities are crucial for creative applications, where multiple iterative edits and refinements are often required to achieve the desired narrative or visual composition. Our method can also perform stylistic transformations, allowing users to apply global changes across all story panels, such as rendering them in the style of `Van Gogh' or `Day of the Dead' (Fig. \ref{fig:qual} (c)). This flexibility is particularly valuable in the creative process: users can initially focus on designing the story's characters and elements, then experiment with different artistic styles during later stages of refinement. Additionally, our method supports personalization. Users can provide a single reference image of a desired concept or a LoRA model, enabling  integration of custom elements into the story (Fig. \ref{fig:qual} (d)). The original versions of these personalized stories can be found in the Appendix.

Furthermore, our method can seamlessly edit user-provided story frames, whether from existing books or generated by other methods, without requiring any special adjustments. To demonstrate its ability to modify existing visual stories, we edited frames from two different stories sourced from the Gutenberg\footnote{\url{https://www.gutenberg.org}} in Fig~\ref{fig:existing}. Our approach successfully transforms characters, such as changing duck to a rooster (Fig. \ref{fig:existing} left), as well as performs fine-grained edits like changing a coat to a trench coat or altering hair color and style (Fig. \ref{fig:existing} right) consistently.  These results highlight the robustness and versatility of our approach, showcasing its ability to perform detailed edits, handle diverse stylistic changes, and support personalized storytelling; essential for enhancing creative workflows.  Please see the Appendix for additional story visualizations, including longer stories.

\begin{table*}[!htp]
    \centering
        \caption{Quantitative results for story generation (first part of the table) and editing tasks (second part) across different methods, evaluated using CLIP-I, CLIP-T, DINO, and LPIPS metrics, along with user study results (1-5 scale) for image consistency (User-I), text alignment (User-T), and disentanglement (User-D).}
     \resizebox{0.80\textwidth}{!}{%
    \begin{tabular}{c lcccccccc}
        \toprule
        & \textbf{Method} & \textbf{CLIP-I}$\uparrow$ & \textbf{CLIP-T}$\uparrow$ & \textbf{DINO}$\uparrow$ &  \textbf{LPIPS}$\downarrow$ & \textbf{User-I}$\uparrow$ & \textbf{User-T}$\uparrow$ & \textbf{User-D} $\uparrow$ & \textbf{Time(s)}$\downarrow$\\
        \midrule
        \multirow{5}{*}{\rotatebox[origin=c]{90}{\shortstack{\textbf{Story} \\ \textbf{Generation}}}}

        & StoryDiffusion~\cite{zhou2024storydiffusion} & 0.78 $\pm$ 0.05 & 0.32 $\pm$ 0.05 & 0.41 $\pm$ 0.10 & 0.49 $\pm$ 0.04 & 2.55$\pm$1.29 & 2.44$\pm$1.27 & N/A & \textbf{9s}\\
        & Intelligent Grimm~\cite{liu2024intelligent} & 0.78 $\pm$ 0.05 & 0.30 $\pm$ 0.02 & \textbf{0.52 $\pm$ 0.11} & 0.63$\pm$0.03 & 2.09$\pm$1.13 & 2.08$\pm$1.16 & N/A & 15s\\
        & Consistory~\cite{tewel2024consistory} & \textbf{0.83 $\pm$ 0.04} & 0.34 $\pm$ 0.02 & 0.49 $\pm$ 0.11 & 0.51 $\pm$0.03 & 2.70$\pm$1.47 & 2.64$\pm$1.52 & N/A & 17s \\
        & AutoStudio~\cite{cheng2024autostudio} & 0.78 $\pm$ 0.05 & 0.33 $\pm$ 0.02 & 0.38 $\pm$ 0.18 & 0.48$\pm$0.18 & 2.73$\pm$1.29 & 2.73$\pm$1.30 & N/A & 11s\\
        & Plot'n Polish (Ours) & 0.79$\pm$0.05 & \textbf{0.36$\pm$0.02} & 0.41 $\pm$ 0.12 &\textbf{0.47$\pm$0.06} & \textbf{2.90$\pm$1.30} & \textbf{2.82$\pm$1.30} & N/A & 11s\\
        \midrule
        \midrule
        \multirow{4}{*}{\rotatebox[origin=c]{90}{\shortstack{\textbf{Story} \\ \textbf{Editing}}}}
  
        & InstructPix2Pix~\cite{instructpix2pix} & 0.89 $\pm$ 0.12 & 0.31 $\pm$ 0.08 & 0.77 $\pm$ 0.24 & 0.21 $\pm$ 0.19 & 2.37 $\pm$ 1.26  & 2.23 $\pm$ 1.21 & 1.65 $\pm$ 0.98 & 9s \\
        & LEDITS++~\cite{brack2024ledits} & 0.76 $\pm$ 0.07 & 0.29 $\pm$ 0.06 & 0.50 $\pm$ 0.16 & 0.24 $\pm$ 0.05 &2.35 $\pm$ 1.21 & 2.62 $\pm$ 1.32 & 2.49 $\pm$ 1.35 & \textbf{5s}  \\
        & Plug-and-Play~\cite{Tumanyan_2023_CVPR} & 0.82 $\pm$ 0.09 & 0.30 $\pm$ 0.06 & 0.67 $\pm$ 0.16 & 0.30 $\pm$ 0.08 & 2.13 $\pm$ 1.21 & 2.41 $\pm$ 1.30 & 2.46 $\pm$ 1.25 & 250s  \\
        & Plot'n Polish (Ours) & \textbf{0.93 $\pm$ 0.06} & \textbf{0.33 $\pm$ 0.07} & \textbf{0.88$\pm$0.17} & \textbf{0.10 $\pm$ 0.07} & \textbf{3.99 $\pm$ 1.07} & \textbf{4.08 $\pm$ 1.06} & \textbf{4.17 $\pm$ 1.04} & 9s \\
        \bottomrule 
    \end{tabular}
     }
        \label{tab:quantitative_results}
\end{table*}

\noindent \textbf{Qualitative Comparison.} We compare our method to both state-of-the-art story visualization methods~\cite{zhou2024storydiffusion, tewel2024consistory, cheng2024autostudio, liu2024intelligent} and editing methods~\cite{instructpix2pix,brack2024ledits,Tumanyan_2023_CVPR}. Note that we were not able to compare with~\cite{gong2023talecrafter, tao2024storyimager} since their code not publicly available. We first evaluate story visualization consistency in Fig. \ref{fig:qual-comp}. Our method excels at maintaining consistent visual elements, such as preserving both girl and Captain characters' attire across all panels. In contrast, other methods struggle to achieve the same level of consistency or accurate character depiction. For instance, StoryDiffusion ~\cite{zhou2024storydiffusion} fails to maintain consistent attire for both characters(see Captain character). Consistory~\cite{tewel2024consistory} maintains some level of consistency but omits the Captain character in the third and sixth panels while introducing additional characters in the first and fifth panels, deviating from the intended narrative. AutoStudio ~\cite{cheng2024autostudio}   frequently changes character attire and generating blended visuals (e.g. Captain is wearing a skirt). Finally, IntelligentGrimm ~\cite{liu2024intelligent} is heavily constrained by its reliance on a specific dataset, limiting the style and flexibility. Additionally, ~\cite{liu2024intelligent} often fails to depict key story elements accurately (e.g. omitting the Captain character entirely in the second panel). In contrast, our method consistently maintains coherent visuals across all panels. See Appendix for more qualitative comparisons.

For editing tasks, we compare our method against both story generation methods that offer some form of editing \cite{zhou2024storydiffusion, cheng2024autostudio} as well as state-of-the-art editing methods.  Fig. \ref{fig:edit-comp} shows a comparison with story generation methods on editing task. Although the generation code for AutoStudio~\cite{cheng2024autostudio} is publicly available, the editing code is not, preventing a quantitative comparison.  However, Fig. \ref{fig:edit-comp} shows a qualitative comparison between their method and ours, using images taken from the original paper. The first 4 panels in each row showcase generated panels, while the last three panels illustrate scenarios where the models are tasked with applying specific edits, such as depicting the character at the beach, adding a dolphin to the sea, or giving the character sunglasses. StoryDiffusion struggles significantly with these editing tasks. For instance, while the character was depicted at the beach as requested, sunglasses were inadvertently added despite not being part of the prompt. Additionally, when prompted to add a dolphin, the model produced an incorrect result by blending the dolphin with the character, creating an unrealistic and unintended visualization. In contrast, both our method and AutoStudio demonstrate the capability to perform multi-turn edits, where successive prompts guide iterative adjustments. However, when AutoStudio  was prompted to depict the character sunbathing, it instead generates a pose where the character is posing to the camera, misinterpreting the context. In comparison, our method accurately captures the intent of the prompt, presenting the character in a realistic sunbathing posture. 

Furthermore, in Fig.~\ref{fig:edit_comp}, we compare our method against state-of-the-art editing methods \cite{instructpix2pix,brack2024ledits,Tumanyan_2023_CVPR}. Each row represents a different edit scenario, including modifications like replacing a butterfly with an eagle or a flower and transforming a dog into a tiger or wolf. InstructPix2Pix ~\cite{instructpix2pix} introduces unintended changes beyond the target object, altering key aspects of the scene, such as modifying the girl's appearance instead of focusing solely on the intended edit. Ledits++~\cite{brack2024ledits} struggles with blending, e.g., merging the dog and tiger unnaturally or distorting the girl's face during transformation. P2P ~\cite{Tumanyan_2023_CVPR} fails to generate the requested objects in some cases, such as being unable to synthesize an eagle or a flower as intended. In contrast, our method maintains both visual consistency (e.g., generating the exact same flower across frames) and semantic accuracy,  and only the specified modifications are applied while keeping the rest of the scene unchanged.

\subsection{Quantitative Experiments}
We conducted quantitative experiments by generating 200 stories, each consisting of 9 frames, resulting in a total of 1800 frames. The initial stories were generated using GPT-4~\cite{achiam2023gpt}, featuring diverse characters and settings. Table \ref{tab:quantitative_results} reports several key metrics: image similarity (CLIP-I~\cite{radford2021learning}, DINO~\cite{caron2021emerging}, LPIPS~\cite{zhang2018perceptual}) and text similarity (CLIP-T~\cite{radford2021learning}). See Appendix for experiment details.

\noindent \textbf{Story Visualization} For story visualization, our method outperforms other approaches in CLIP-T, indicating its superior ability to generate visuals that align closely with the provided text prompts. Additionally, our method achieves higher CLIP-I scores than most competitors, demonstrating its ability to maintain consistency across story panels—a crucial aspect of coherent storytelling. An exception is ConsiStory~\cite{tewel2024consistory}, which attains a high CLIP-I score of 0.83. However, this comes at a significant cost: as shown in Fig.~\ref{fig:qual-comp}, ConsiStory's generated panels fail to accurately depict the intended characters, either omitting them or introducing additional ones (see Appendix for more examples), diverging from the intended narrative. Furthermore, our method outperforms in LPIPS and DINO metrics, reinforcing that we generate consistent story panels.

\noindent \textbf{Story Editing}  For editing task (see Table \ref{tab:quantitative_results}) our method outperforms  in both preserving the original image and aligning with the given edit prompt. It achieves the highest CLIP-I  and DINO  scores, indicating strong consistency with the original image, and the lowest LPIPS, demonstrating minimal unintended changes. Additionally, it achieves the best CLIP-T scores,   confirming our method’s superior ability to apply edits while respecting to the original image.

\noindent \textbf{User Study.} We employed  
 50 participants from Prolific.com across 60 story frames for each method for user assessment. Each participant was shown story panels where the corresponding narrative was included, and asked to evaluate two key aspects of the story visualization: 1)  Alignment with the narrative 2) Consistency across panels. Table \ref{tab:quantitative_results} presents the results, comparing our method to state-of-the-art approaches. Our method was rated highest for both narrative alignment and consistency, demonstrating its ability to generate visuals that closely follow the story text while maintaining coherent character and object depictions.  For editing, each participant was presented with three images alongside their corresponding edits and asked to assess three key aspects: 1) Alignment with the edit prompt , 2) Disentanglement and 3) Consistency. For both user studies, we used a rating scale from 1 (Very Bad) to 5 (Very Good). Our method received the highest ratings across all aspects, demonstrating its ability to perform consistent and disentangled editing. See Appendix for more details.
 
\begin{figure*}
    \centering
    \includegraphics[width=0.8\textwidth]{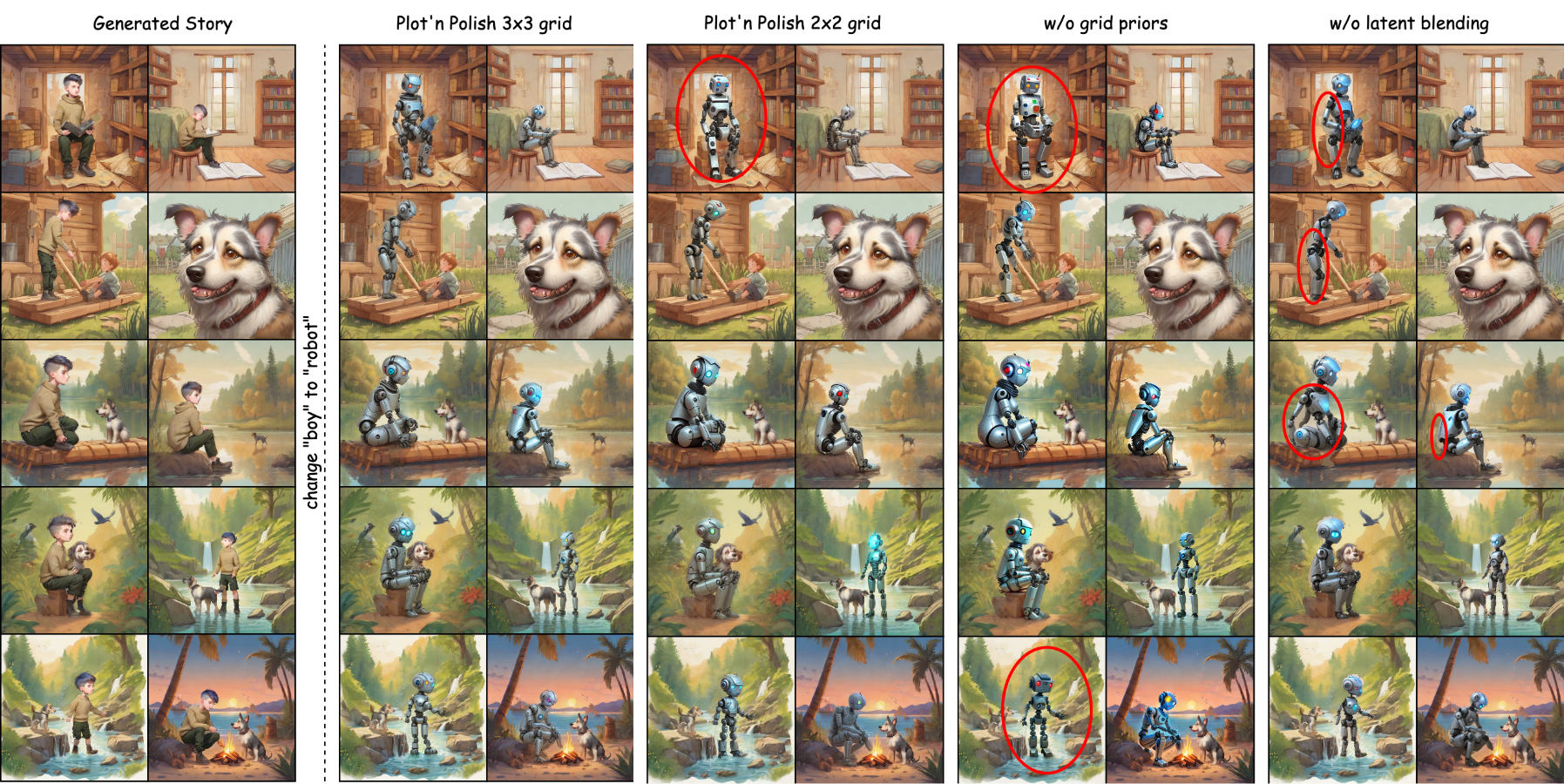}
    \caption{We ablate the following components: grid size, grid priors, and latent blending. See red-circled areas for inconsistencies in characters and background artifacts.}
    \label{fig:ablation}
\end{figure*}

\begin{figure}
    \centering
    \includegraphics[width=0.40\textwidth]{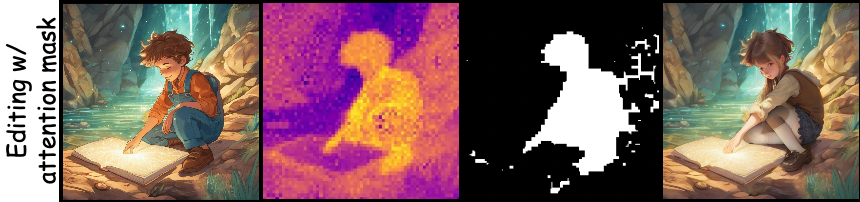}
    \caption{Example of image editing using attention-derived mask.}
    \label{fig:attn-mask}
\end{figure}

\noindent \textbf{Story Generation and Editing Time.} 
Our method exhibits high efficiency, generating initial frames in just 2 seconds and an editing speed of 9 s per frame, totaling 11 s. As shown in Table \ref{tab:quantitative_results}, our approach performs comparably to the state-of-the-art story generation and editing methods while maintaining a balance between speed and quality.

\noindent \textbf{Ablation Study.}
Our ablation study investigates the impact of different consistency mechanisms on the stories generated, as illustrated in Figure~\ref{fig:ablation} (see an additional example in the appendix). Increasing the size of the grid enhances the frame-to-frame interaction, which, in turn, improves visual consistency. For instance, the 3×3 grid exhibits better character continuity compared to the 2×2 variant. When grid priors are removed, meaning that each frame is edited independently, the model lacks any explicit mechanism for enforcing consistency, often resulting in noticeable variations in character appearance. Likewise, omitting latent blending in favor of a copy-and-paste approach introduces visible artifacts between frames. These findings underscore the importance of both grid structure and latent blending in maintaining coherent visual narratives.

Furthermore, our method is inherently flexible and can accommodate alternative masking strategies. To test this flexibility, we performed an additional experiment using attention-based masks derived from the model’s internal attention maps, replacing explicit segmentation. As shown in Figure~\ref{fig:attn-mask}, while segmentation masks offer more precise and controllable edits, attention-derived masks still yield compelling results. This demonstrates the robustness of our approach in different sources of masks.

We also verify that our method generalizes effectively to high-resolution diffusion models such as SDXL~\cite{podell2023sdxlimprovinglatentdiffusion} and Flux~\cite{flux2024}, with results provided in the Appendix.
\section{Limitation and Societal Impact}
\label{sec:limitation}
\begin{figure}
    \centering
    \includegraphics[width=0.45\linewidth]{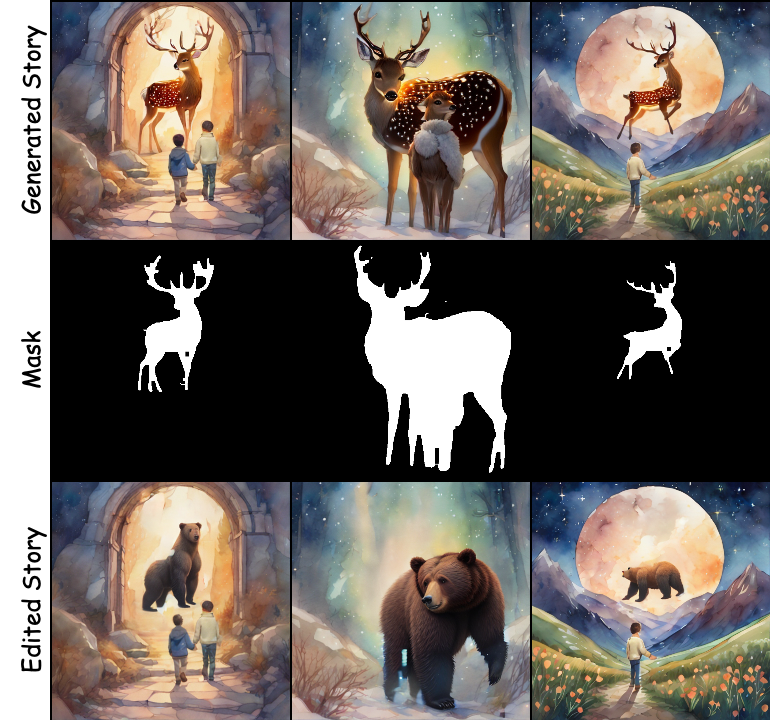}
    \caption{Failure case: Due to overlapping mask detection, the middle figure incorrectly includes additional subjects.}
    \label{fig:limitation}
\end{figure}

Our method leverages segmentation models to detect masks for characters or objects targeted for editing. As a result, the quality of our edits is influenced by the performance of these models. For instance, in cases where multiple overlapping objects exist, such as two reindeer in Fig.~\ref{fig:limitation}, our method may edit the unintended subjects. Although our method relies on a segmentation model such as \cite{xiong2023efficientsamleveragedmaskedimage}, it can efficiently process an entire batch of 10 images in just 2 seconds, making it a more practical and scalable solution.

\section{Conclusion}
\label{sec:conclusion}
In this paper, we introduce Plot'n Polish, a novel framework for story visualization that overcomes key limitations of existing methods by allowing multi-frame editing, ensuring consistency, and supporting fine-grained and large-scale modifications. Unlike prior approaches, our method allows users to edit entire story sequences while preserving coherence, making changes such as altering attire, adding accessories, replacing characters, or transforming visual styles. Edits are applied consistently across frames without requiring full regeneration, enhancing creative control. Additionally, personalization features, including user-provided concepts and styles, further expand its versatility. Future work will explore interactive branching narratives to support more complex story dynamics.
{
    \small
    \bibliographystyle{ieeenat_fullname}
    \bibliography{main}
}

\newpage
\setcounter{secnumdepth}{1}
\setcounter{section}{0}  % Reset section counter
\renewcommand{\thesection}{\Alph{section}}  % Use letters instead of numbers
\twocolumn[{%
\renewcommand\twocolumn[1][]{#1}%
\centering
\includegraphics[width=.80\linewidth]{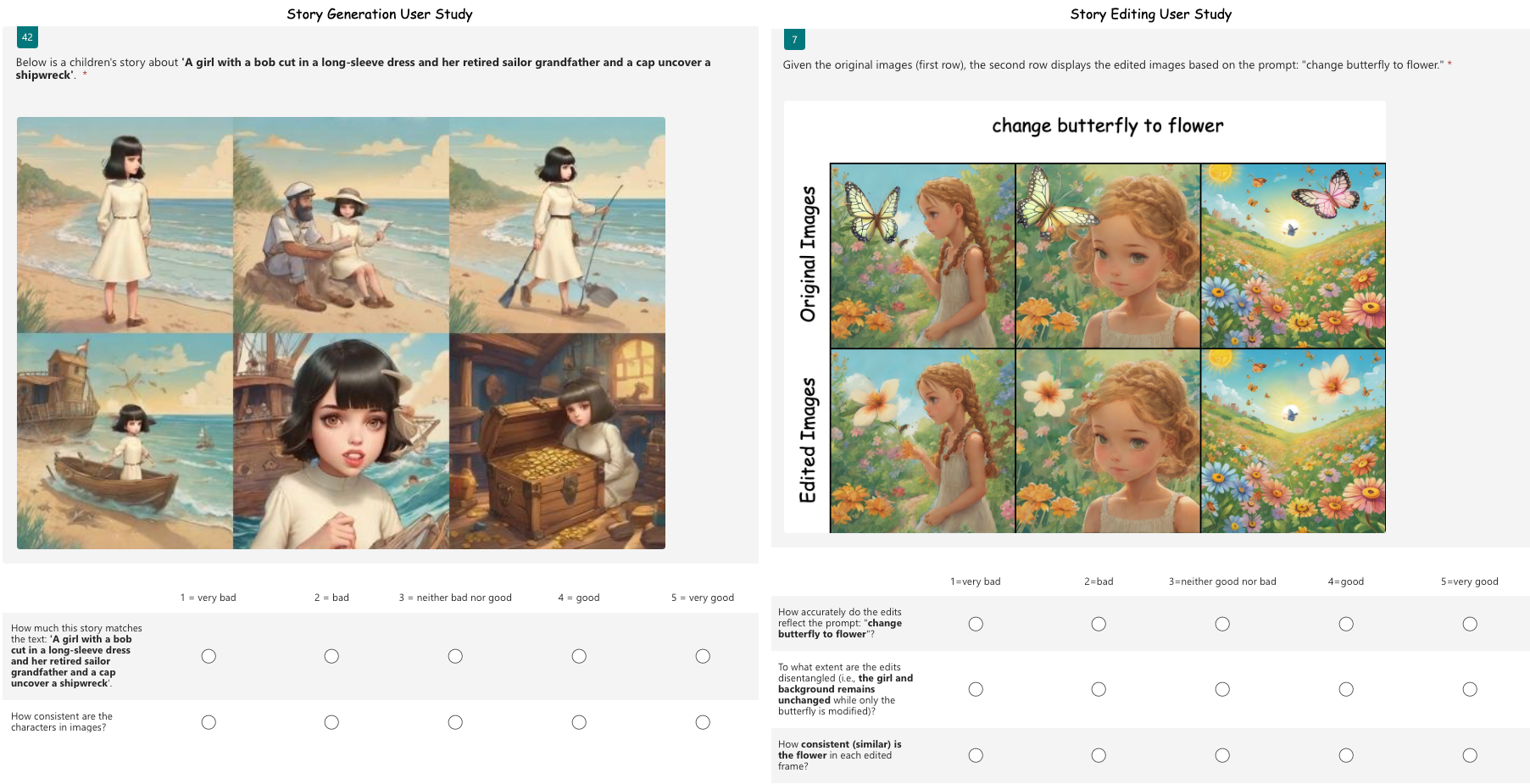}
\captionof{figure}{Screenshot from user study evaluating consistency, alignment, and disentanglement.}
\vspace{1em}
\label{fig:user-study}
}]

\section{Story Generation with LLMs}
\label{sec:supp-a}
Given an initial idea $\mathcal{I}$ such as 'A story about a boy and his toy truck', the desired number of story pages $n$, the story style $\mathcal{S}_{\text{story}}$ such as 'children's picture book style' or 'comic book style', and the visualization style $\mathcal{S}_{\text{visual}}$ such as `watercolor painting style', we employ GPT-4~\cite{achiam2023gpt} to produce a set of image prompts and corresponding story plot text that align with the provided inputs. Our prompt engineering ensures that the LLM produces visually descriptive prompts where story elements remain consistent throughout, unnecessary details like reasons for actions or dialogue are omitted, and misleading details are avoided (see Supplementary Material for details). The generation process is formalized as:
\[
(\mathcal{P}, \mathcal{T}) = \left\{ \mathcal{P}_i, \mathcal{T}_i \right\}_{i=1}^n = \mathrm{LLM}\left( \mathcal{I}, n, \mathcal{S}_{\text{story}}, \mathcal{S}_{\text{visual}} \right)
\]
where $T_i$ represents the plot text and $P_i$ represents the prompt for the i-th story page. Each prompt $P_i$ consists of three key components:
\begin{equation}
    P_i = P_{context} + P_{background} + P_{style}
\end{equation}
where $P_{context}$ contains descriptions of characters and their actions within the scene, $P_{background}$ provides detailed descriptions of the background elements, and $P_{style}$ incorporates the specified visualization style as input by the user. We achieve background and style consistency with these detailed prompts.

Our pipeline presents LLM-generated narratives to users in a structured JSON format (see Listing \ref{lst:json} for an example). Our story visualization and refinement component, detailed in the next section, can directly utilize this JSON file to create visualizations or modify objects and styles based on user-provided text prompts. However, if desired, users have the option to review and refine these prompts and plots before proceeding to visualization. Each iteration of feedback involves reintroducing the conversation history between the LLM and the user back into the model. This method allows the LLM to make targeted adjustments based on user feedback, ensuring consistency in other parts of the story. The process continues iteratively until the user is satisfied with the story development.

\section{Further Details}
\label{sec:supp-results}
\paragraph{User Study.} 
 Each participant was shown story panels where the corresponding narrative was included and asked to evaluate two key aspects of the story visualization. 1)  Alignment with the narrative: `How well do the visuals capture the story narrative described by the text?' 2) Consistency across panels: `How consistent are the characters and objects across the story panels?'. 
 
 For editing evaluation, each participant was presented with three images alongside their corresponding edits and asked to assess three key aspects: 1)Alignment with the edit prompt: How accurately do the edits reflect the prompt? 2) Disentanglement: To what extent are the edits disentangled? 3) Consistency: How consistent (similar) is the subject in each edited frame? For both user studies, we used a rating scale from 1 (Very Bad) to 5 (Very Good). Our method received the highest ratings across all aspects, demonstrating its ability to perform consistent and disentangled editing. Figure~\ref{fig:user-study} presents screenshots from both of our user studies.

\onecolumn
\begin{lstlisting}[language=json, caption={JSON output generated by the LLM (see Section~\ref{sec:supp-a}) given the idea "A scientist woman creates a weather-changing machine".}, label={lst:json}]
{
    "Main Characters": [
        {
            "Name": "Dr. Mira",
            "Description": "Woman with hair pulled back, lab coat",
            "Category": "woman"
        },
        {
            "Name": "Robin",
            "Description": "Sparrow with a tuft on the head",
            "Category": "sparrow"
        }
    ],
    "Story": [
        {
            "Page": 1,
            "Text": "Dr. Mira was fascinated by the weather and was determined to create a machine that could change it with just the push of a button.",
            "Image_Prompt": "Dr. Mira standing in her lab beside a large, intricate weather-changing machine with blinking lights and rotating gears",
            "Location_Description": "A cluttered laboratory filled with various scientific instruments and a large window overlooking the park",
        },
        {
            "Page": 2,
            "Text": "After many trials and errors, her machine was finally ready. She activated it with a smile, completely unaware of the consequences.",
            "Image_Prompt": "Dr. Mira pushing a large red button on the machine, looking excited",
            "Location_Description": "cluttered laboratory",
        },
        {
            "Page": 3,
            "Text": "The machine roared to life and, to Dr. Mira's delight, the sunny skies turned cloudy and a gentle rain began to pour.",
            "Image_Prompt": "Close-up of Dr. Mira's face, her expression a mix of awe and success",
            "Location_Description": "room",
        },
        {
            "Page": 4,
            "Text": "However, the gentle rain soon turned into a torrential downpour, flooding the local park where a little sparrow, Robin, was taking shelter.",
            "Image_Prompt": "Robin the sparrow struggling against the heavy rain, perched on a branch",
            "Location_Description": "A vibrant local park with benches, trees, and a small pond",
        },
        {
            "Page": 5,
            "Text": "Seeing the chaos she had inadvertently caused, Dr. Mira rushed to adjust the settings on her machine.",
            "Image_Prompt": "Dr. Mira hastily tweaking dials and pressing buttons on the machine with a worried expression",
            "Location_Description": "cluttered laboratory",
        },
        {
            "Page": 6,
            "Text": "The weather calmed and Robin flew to Dr. Mira's window, chirping gratefully. Dr. Mira realized she needed to consider the impact of her experiments.",
            "Image_Prompt": "Robin perched on Dr. Mira's open window, chirping",
            "Location_Description": "A view from the laboratory with the window open towards the sunny park",
        }
    ]
}
\end{lstlisting}
\twocolumn

\paragraph{Quantitative Experiments}
We generated 200 unique story ideas using GPT-4, each featuring two characters and generated prompts as described in Section~\ref{sec:supp-a}. Each story consists of 9 frames, resulting in a total of 2000 images.  

To evaluate the effectiveness of our editing approach, we applied 100 edits per image, amounting to 180,000 total edits across all stories. Among these, we prompted GPT-4 to transform characters into 30 different animals and 30 objects while also performing 10 clothing color changes and 10 hair color changes per story. Additionally, for global edits, we applied 20 distinct visual styles.

\section{Extended Experiments and Ablations}

\begin{figure}
    \centering
    \includegraphics[width=\linewidth]{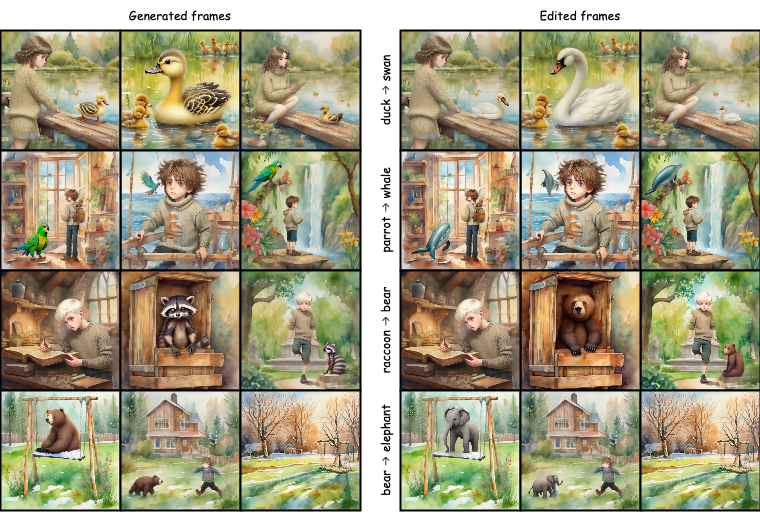}
    \caption{Further qualitative examples of our editing method.}
    \label{fig:edit-supp}
\end{figure}

\textbf{Additional Qualitative Editing Results.}  
Figure~\ref{fig:edit-supp} presents additional qualitative results of our consistent editing method. These examples further demonstrate the ability of our approach to make precise, localized changes while maintaining visual coherence across story frames.

\begin{figure}[h]
    \centering
    \includegraphics[width=\linewidth]{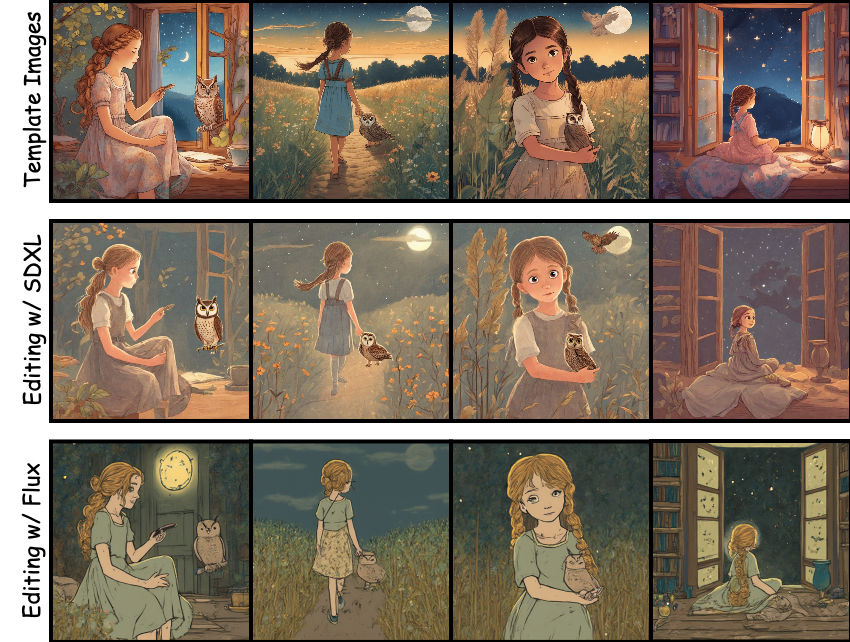}
    \caption{Editing results using SDXL and Flux.}
    \label{fig:edit-xl}
\end{figure}

\textbf{Editing with SDXL and Flux.}  
Our method generalizes effectively to models designed for high-resolution synthesis, such as SDXL~\cite{podell2023sdxlimprovinglatentdiffusion} and Flux~\cite{flux2024}. In these cases, each frame is generated at 1024×1024 resolution, preserving both visual quality and coherence across the full composition (see Figure~\ref{fig:edit-xl}).

\begin{figure}
    \centering
    \includegraphics[width=\linewidth]{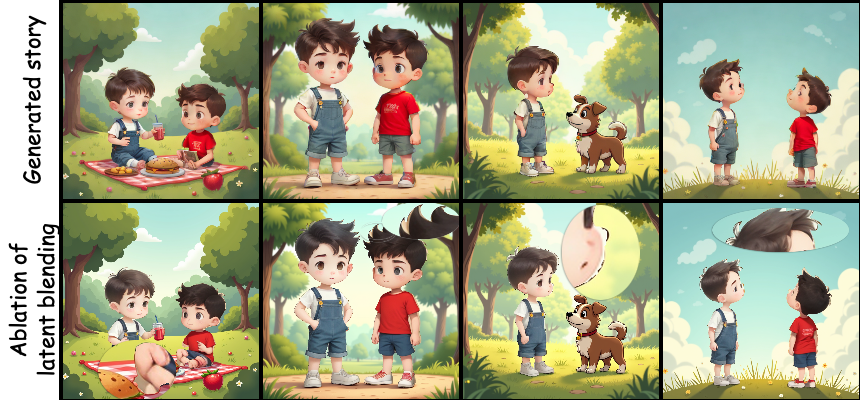}
    \caption{Ablation example showing artifacts caused by the absence of latent blending. Zoomed-in views reveal unnatural transitions and inconsistencies, underscoring the importance of our blending strategy for spatial coherence.}
    \label{fig:ablation-supp}
\end{figure}

\textbf{Ablation Study on Latent Blending.}  
To provide further insight into the effects of latent blending, we include an additional example that highlights the presence of unnatural artifacts when blending is omitted. Zoomed-in views are provided to enhance clarity (see Figure~\ref{fig:ablation-supp}).

For instance, in the second frame of the bottom row, omitting latent blending results in distorted hair and facial artifacts that disrupt character identity. In the third frame, a large, disjointed close-up of the character’s face appears unnaturally overlaid on the scene, breaking narrative continuity. Additionally, the first frame shows misaligned limbs and unnatural character positioning during the picnic scene. These artifacts are not present in the corresponding top-row frames, which were generated using our blending strategy. Such comparisons underscore the importance of latent blending for maintaining spatial and semantic coherence across story frames.

\begin{figure}
    \centering
    \includegraphics[width=\linewidth]{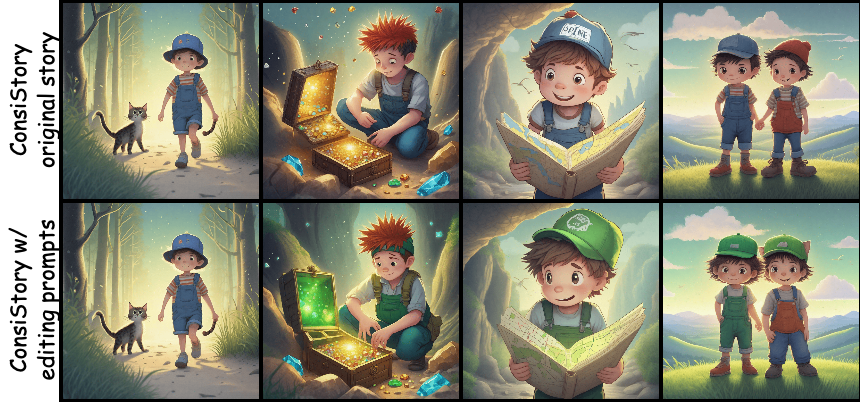}
    \caption{Example of prompt editing using ConsiStory: replacing “cap” with “green hat” results in unintended global changes—such as altered facial expressions, clothing, and edits to unrelated areas. This highlights the limitations of prompt-level editing for achieving precise, localized control.}
    \label{fig:edit-prompts}
\end{figure}

\begin{figure*}
    \centering
    \includegraphics[width=\linewidth]{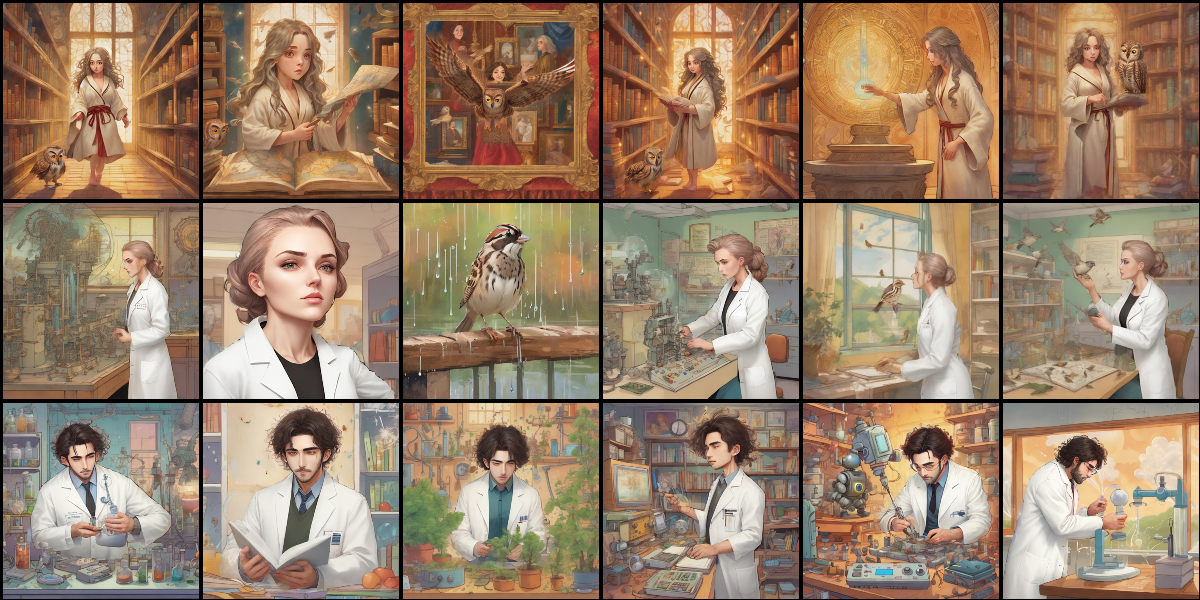}
    \caption{Original versions of edited stories, generated with text prompts such as "a girl with a robe"; personalized versions are shown in Figure~\ref{fig:qual} (d).}
    \label{fig:pers_orig}
\end{figure*}

\textbf{Limitations of Prompt-Based Editing.}  
Story generation models are not inherently designed for precise local modifications. As shown in Figure~\ref{fig:edit-prompts}, altering prompts—for example, replacing ``cap'' with ``green hat''—often leads to unintended global changes: modified facial expressions (3rd frame), clothing inconsistencies (e.g., overalls in the 2nd, 3rd, and 4th frames), or unrelated alterations (e.g., the treasure in the 2nd frame). These artifacts illustrate the limitations of prompt-only editing approaches, which lack mechanisms for spatial or semantic control.
Such results reinforce the need for a post-hoc editing framework—like ours—that enables localized and consistent changes without compromising surrounding narrative elements.
\textbf{Personalized Editing from Image References.}  
See Figure~\ref{fig:pers_orig} for the original versions of the edited stories, generated using text prompts such as ''a girl with a robe.'' The corresponding personalized results, generated using image references, are shown in Figure~\ref{fig:qual}(d).

\begin{figure*}
    \centering
    \includegraphics[width=\linewidth]{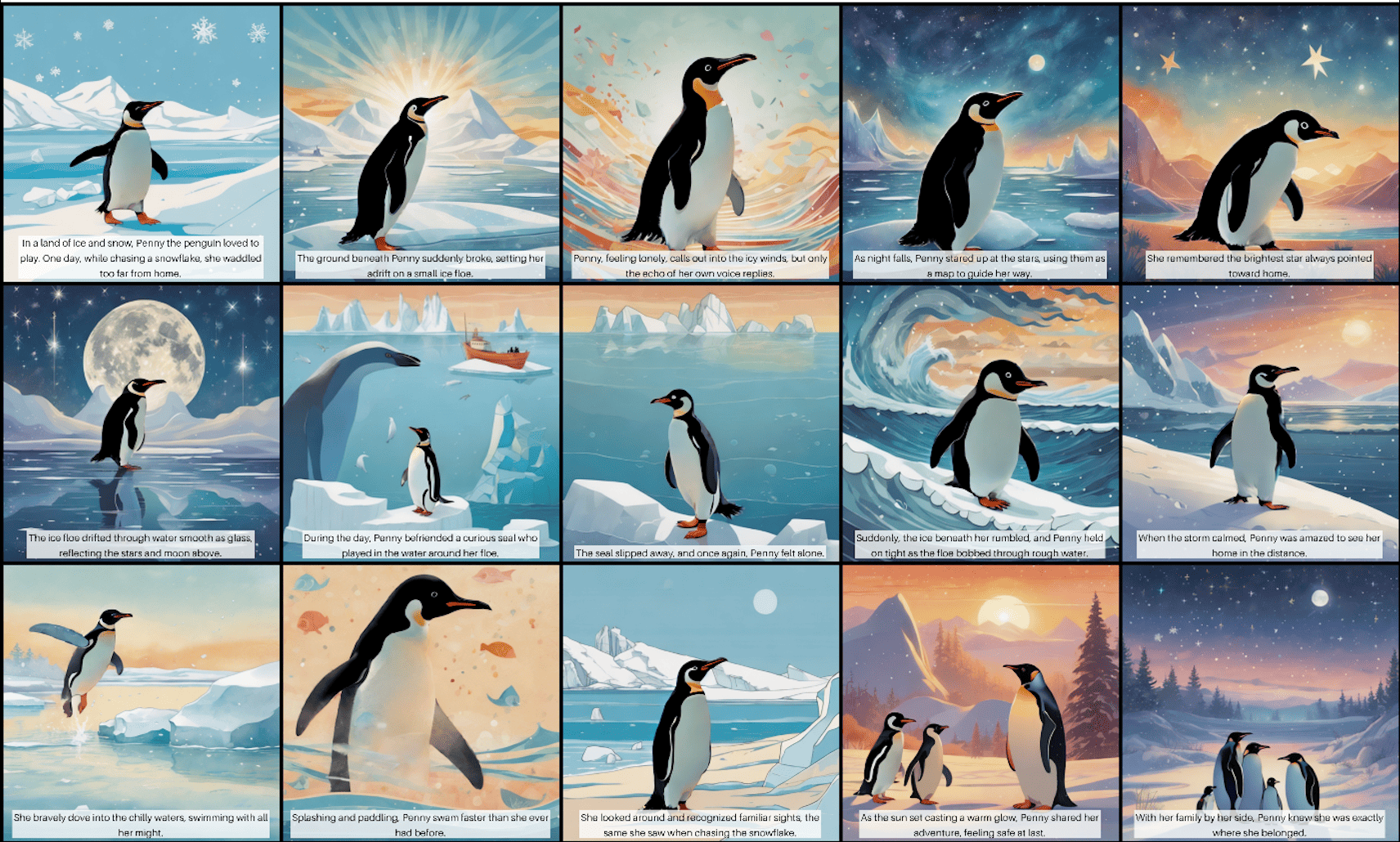}
    \caption{Our method is capable of generating longer stories featuring a diverse range of characters.}
    \label{fig:longer}
\end{figure*}

\begin{figure*}
    \centering
    \includegraphics[width=0.95\linewidth]{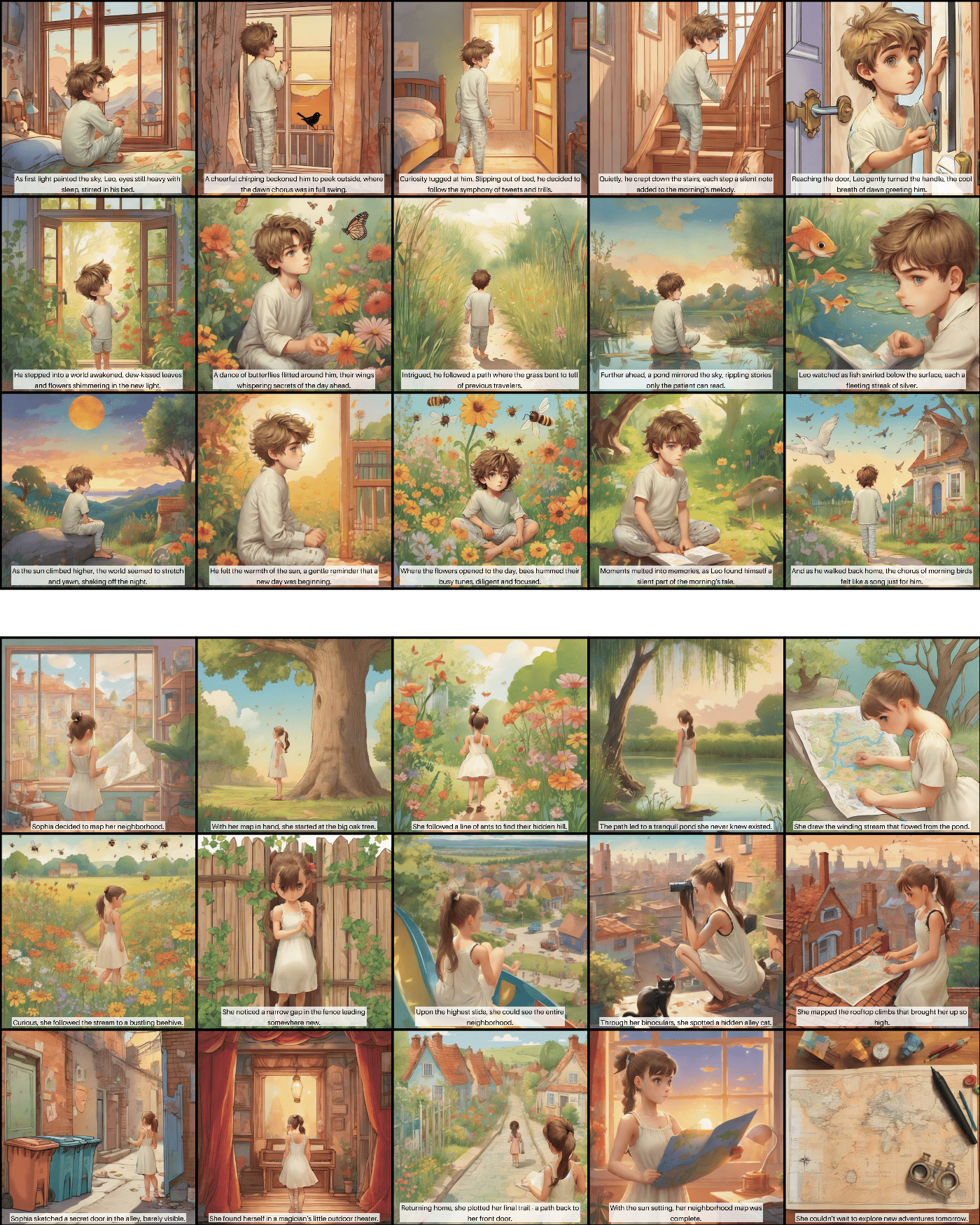}
    \caption{Our method can generate long stories with a diverse range of characters, capturing different attributes and ensuring consistency across the narrative.}
    \label{fig:longer2}
\end{figure*}

\textbf{Scalability to Longer and Diverse Stories.}  
Figures~\ref{fig:longer} and~\ref{fig:longer2} showcase longer stories generated by our method, demonstrating its scalability. Our approach effectively handles extended narratives and supports a wide range of characters, including both animals and humans, while preserving consistency and visual quality throughout.

\section{Extended Qualitative Comparisons}

\textbf{Comparison with Video Editing Methods.}  
While story editing shares structural similarities with video editing—namely, the need for coherence across frames—key differences exist. Story frames often exhibit greater semantic variation than temporally adjacent video frames. Furthermore, most video editing methods are designed for global modifications and struggle with fine-grained, localized edits.

\begin{figure}
    \centering
    \includegraphics[width=\linewidth]{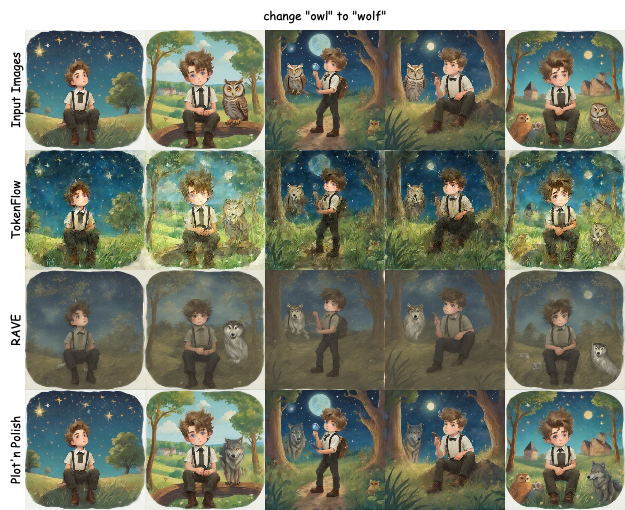}
    \caption{Comparison with video editing methods.}
    \label{fig:video}
\end{figure}

To evaluate this, we compare our approach with two state-of-the-art video editing methods: TokenFlow~\cite{tokenflow2023} and RAVE~\cite{kara2024rave} (Figure~\ref{fig:video}). TokenFlow fails to apply the requested transformation (e.g., changing an owl to a wolf), while RAVE produces overextended modifications, impacting the entire scene. In contrast, our method performs precise, localized edits while preserving context and visual integrity across the story frames. This illustrates our approach’s advantage in scenarios where detailed and consistent modifications are essential.]

\begin{figure*}[t!]
    \centering
    \includegraphics[width=0.75\linewidth]{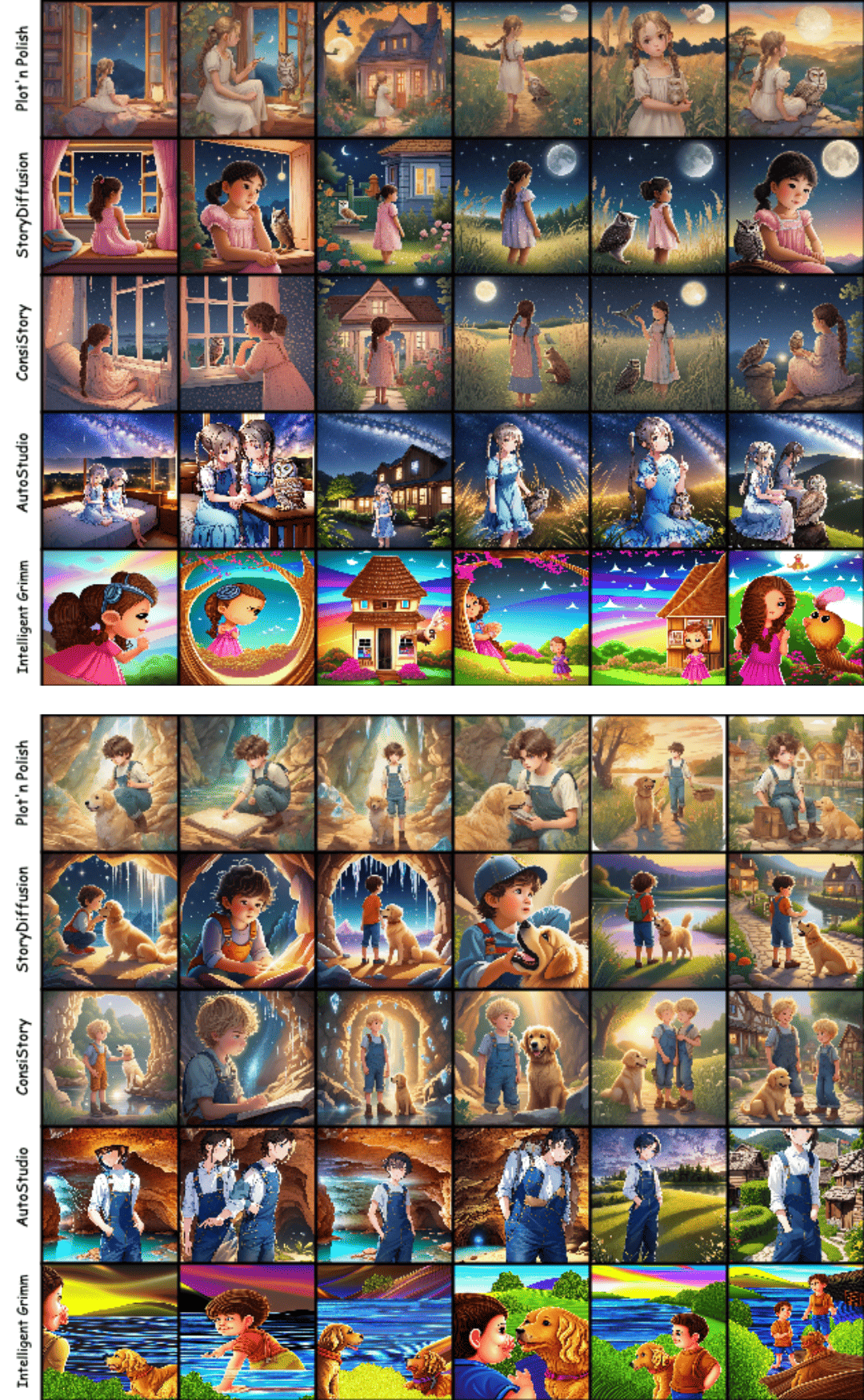}
    \caption{Further qualitative comparisons of our method against state-of-the-art story visualization techniques, including StoryDiffusion, ConsiStory, AutoStudio, and Intelligent Grimm. }
    \label{fig:qual-comp-supp}
\end{figure*}

\textbf{Additional Comparisons for Story Generation.}  
Figure~\ref{fig:qual-comp-supp} provides further qualitative comparisons of story generation outputs. These results support findings in Section~\ref{sec:exp}, illustrating limitations of prior methods. AutoStudio~\cite{cheng2024autostudio} frequently fails to align with input text, instead generating redundant or duplicated characters. Intelligent Grimm~\cite{liu2024intelligent} struggles to produce coherent visuals for novel scenarios. While StoryDiffusion~\cite{zhou2024storydiffusion} achieves reasonable text alignment, it fails to maintain consistent character appearances (e.g., varying accessories and dress color across frames). Additionally, StoryDiffusion lacks pose diversity, whereas our method supports dynamic and expressive character actions. ConsiStory~\cite{tewel2024consistory} often introduces extra characters or ignores textual cues.

\end{document}